%% file: main.tex
\documentclass[lettersize,journal]{IEEEtran}

\input{preamble}

\begin{document}
\title{Camera Calibration via Circular Patterns:
\\A Comprehensive Framework with Detection Uncertainty and Unbiased Projection Model}

\author{
Chaehyeon Song,
Dongjae Lee,
Jongwoo Lim,
and Ayoung Kim
\thanks{Manuscript received \bl{April 19, 2021}; revised \bl{August 16, 2021.}}
\thanks{This work was supported by the Institute of Information \& communications Technology Planning \& Evaluation (IITP) grant funded by the Korea government(MSIT) No.2022-0-00480}
\thanks{
Chaehyeon song, Dongjae Lee, Jongwoo Lim, and Ayoung Kim are with Seoul National University, Korea (email: chaehyeon, pur22, jongwoo.lim, ayoungk @snu.ac.kr)
}

}

\markboth{IEEE TRANSACTIONS ON PATTERN ANALYSIS AND MACHINE INTELLIGENCE,~Vol.~14, No.~8, August~2025}%
{Shell \MakeLowercase{\textit{et al.}}: A Sample Article Using IEEEtran.cls for IEEE Journals}

\maketitle
\begin{strip}
  \centering
  \vspace{-18ex} 
  \includegraphics[trim=13 4 21 5, clip, width=1\textwidth]{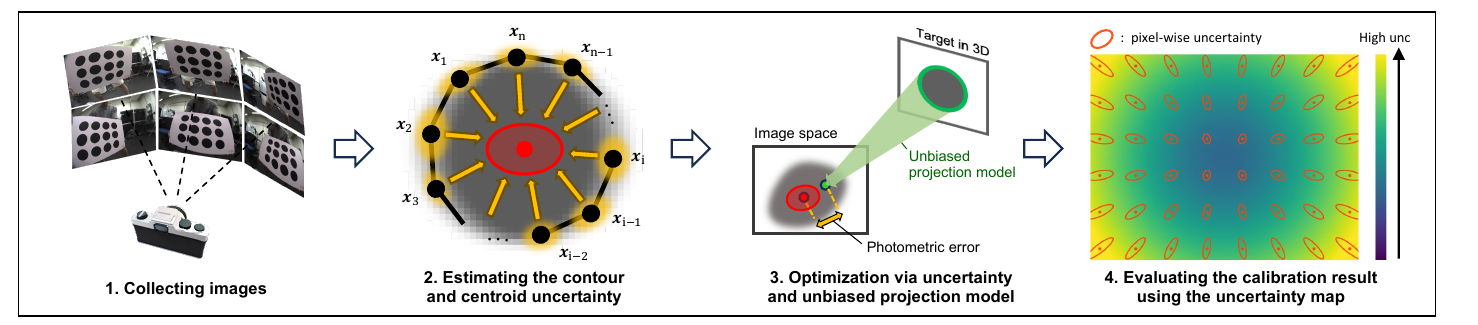}
  \captionof{figure}{\textbf{The uncertainty-aware calibration framework with unbiased projection model.} The core modules---an uncertainty-aware detector and an unbiased estimator---jointly provide both camera calibration parameters and their associated uncertainty for assessing calibration quality.}
  \label{fig:overview}
  \vspace{-2ex} 
\end{strip}
\input{sec/0_abstract} 
\input{sec/1_intro}
\input{sec/2_related_work}

\input{sec/3_Shape_unc}

\input{sec/4_method}
\input{sec/5_results}

\input{sec/6_guide}
\input{sec/7_conclusion}

{
    \small
    \bibliographystyle{IEEEtranN}
    \bibliography{string-long,reference}
}

\input{sec/8_biography}

\input{sec/9_appendix}

\vfill

\end{document}

%% file: preamble.tex
\usepackage[caption=false,font=normalsize,labelfont=sf,textfont=sf]{subfig}
\usepackage{amsmath,amsfonts}
\usepackage{array}
\usepackage{textcomp}
\usepackage{stfloats}
\usepackage{url}
\usepackage{verbatim}
\usepackage{graphicx}
\usepackage[numbers]{natbib}
\usepackage{cuted}       
\usepackage{caption}     
\usepackage[font=footnotesize,labelfont=bf]{caption}
\usepackage{xspace}

\usepackage{soul} 
\usepackage{lipsum}
\usepackage{xcolor}
\definecolor{lightblue}{rgb}{0.8,0.9,1}
\definecolor{lightgreen}{rgb}{0.9,1,0.9}
\definecolor{lightred}{rgb}{1,0.5,0.5}
\definecolor{gray}{rgb}{0.7,0.7,0.7}

\usepackage{multirow}

\usepackage{amsthm}
\usepackage{ifthen}
\usepackage{amsmath}
\usepackage{amssymb}
\usepackage{cancel}

\newtheorem{thm}{Theorem}

\newtheorem{lemma}[thm]{Lemma}

\DeclareMathOperator*{\argmin}{argmin}

\DeclareMathOperator{\cross}{\times}   

\providecommand{\dvec}[1]{\ensuremath{\overleftrightarrow{#1}}}

\usepackage[ruled]{algorithm}
\usepackage{algpseudocode}

\usepackage[amssymb,cdot]{SIunits}
\providecommand{\A}{\ampere}        

\providecommand{\eqoref}[1]{(\ref{#1})}
\providecommand{\eqtref}[2]{(\ref{#1}, \ref{#2})}
\providecommand{\figref}[1]{Fig.~\ref{#1}}
\providecommand{\tabref}[1]{Table~\ref{#1}}
\providecommand{\algoref}[1]{Algorithm~\ref{#1}}
\providecommand{\secref}[1]{\textsection\ref{#1}}

\providecommand{\thmref}[1]{Theorem.~\ref{#1}}
\providecommand{\lemref}[1]{Lemma.~\ref{#1}}

\long\def\symbolfootnote[#1]#2{\begingroup%
\def\thefootnote{\fnsymbol{footnote}}\footnote[#1]{#2}\endgroup}

\newcommand{\bl}[1]{{\textcolor{blue}{#1}}}
\newcommand{\bs}[1]{{\boldsymbol{#1}}}

\newcommand{\eg}{\emph{e.g.,}\xspace}
\newcommand{\ie}{\emph{i.e.}\xspace}

\usepackage{acronym}
\acrodef{SLAM}{simultaneous localization and mapping}
\acrodef{TIR}{thermal infrared}
\acrodef{IoU}{Intersection over Union}

%% file: sec/0_abstract.tex
\begin{abstract}

Camera calibration using planar targets has been widely favored, and two types of control points have been mainly considered as measurements: the corners of the checkerboard and the centroid of circles. Since a centroid is derived from numerous pixels, the circular pattern provides more precise measurements than the checkerboard. However, the existing projection model of circle centroids is biased under lens distortion, resulting in low performance. To surmount this limitation, we propose an unbiased projection model of the circular pattern and demonstrate its superior accuracy compared to the checkerboard.
Complementing this, we introduce uncertainty into circular patterns to enhance calibration robustness and completeness. Defining centroid uncertainty improves the performance of calibration components, including pattern detection, optimization, and evaluation metrics. We also provide guidelines for performing good camera calibration based on the evaluation metric.
The core concept of this approach is to model the boundary points of a two-dimensional shape as a Markov random field, considering its connectivity. The shape distribution is propagated to the centroid uncertainty through an appropriate shape representation based on the Green theorem. Consequently, the resulting framework achieves marked gains in calibration accuracy and robustness. The complete source code and demonstration video are available at \url{https://github.com/chaehyeonsong/discocal}.


\end{abstract}

%% file: sec/1_intro.tex
\section{Introduction}
\label{sec:intro}
A camera projects the 3D world onto a 2D image plane, and camera models mathematically describe this projection using variables called intrinsic parameters. Camera calibration is the process of determining these parameters. Since it enables the unprojection of image pixels to a pencil of 3D rays entering the camera, many 3D computer vision and photogrammetry tasks demand accurate camera calibration, as in 3D dense reconstruction \cite{ECCV-2020-mildenhall, ToG-2022-muller}, visual \ac{SLAM}~\cite{TRO-2015-mur, TRO-2018-qin, RAL-2022-jeon}, and depth estimation~\cite{CVPR-2017-godard}.

The most widely used approach in calibration utilizes the homography characteristic of planar targets~\cite{TPAMI-2000-zhang, TPAMI-2000-heikkila, ICPR-2000-yang, IROS-2013-richardson}, where the dedicated patterns, such as checkerboards~\cite{TPAMI-2000-zhang, CVPR-1999-sturm} or circles~\cite{TPAMI-2000-heikkila, TPAMI-2006-kannala}, are printed. These patterns are used to define control points (\eg corner points of checkerboards or centroids of circular shapes). 
The intrinsic parameters are obtained by minimizing the reprojection error between the locations of the control points in the image and their computed positions based on the camera model. 
Therefore, both precise measurements from the images and the unbiased projection of control points are essential for accurate calibration.

As a centroid is derived from numerous pixels of the two-dimensional shape, the circular pattern provides better measurements than the checkerboard~\cite{ICPR-1996-heikkila} in terms of precision. 
Unfortunately, despite this advantage in measurement, existing circle projections~\cite{ECCV-2004-chen, TPAMI-2005-kim} severely undermine the results by losing the conic property due to lens distortion.
Specifically, the projected position of the original circle center does not correspond to the centroid of a circular shape in the image. 
The errors introduced by this biased projection are more severe than those arising from inaccurate measurements of checkerboard corners~\cite{PAL-2007-mallon, CVPR-2024-song}; therefore, the literature has focused more on improving the checkerboard~\cite{IROS-2013-richardson, ICCV-2019-Peng}. 

A key challenge is that checkerboard-based methods often fail to detect corners or produce unreliable calibration results when image quality is degraded by low resolution, motion blur, or boundary‐blur effects in \ac{TIR} cameras. This limitation prompts us to revisit circular patterns, which offer high-quality measurements despite the mathematical challenges involved. We propose the unbiased projection model of the circular pattern under the pinhole camera model with lens distortion, using probabilistic concept moments~\cite{CVPR-2024-song}. This work addresses the core limitation and demonstrates the superior accuracy of the circular pattern over the checkerboard.
While calibration has been examined from the perspective of accuracy, modeling the uncertainty of calibration results is also vital. In practice, measurements often contain significant noise, and cameras deviate from the ideal camera model~\cite{IJCV-2022-hagemann}. These issues have motivated researchers to define trust regions for calibration results~\cite{IROS-2013-richardson, ICCV-2019-Peng, IJCV-2022-hagemann}, modeling intrinsic parameters not as fixed values but as probabilistic distributions. Such probabilistic modeling enhances the accuracy and robustness of downstream tasks~\cite{CVPR-2024-Xue, CVPR-2021-Zhuang, ICRA-2013-Ozog} and facilitates seeking the next-best-view for camera calibration~\cite{IROS-2013-richardson, ICCV-2019-Peng}. Estimating calibration uncertainty commences with deriving the uncertainty of the measurement. However, related works have only researched the uncertainty of the checkerboard (\ie detection uncertainty of corners), while the uncertainty of the circular pattern has not been addressed.

To introduce uncertainty into the circular pattern, we model the boundary points of a two-dimensional shape as an unidirectional Markov random field. The centroid uncertainty is derived from the covariance of boundary points based on the Green theorem. This approach enables us to consider both the connectivity of boundary points and information from image gradients. 
Our work holds academic value in being the first to present the centroid uncertainty of an arbitrary shape while proving its practical value by proposing the advanced pattern detection method, robust optimization process, and ways to evaluate calibration performance based on this uncertainty as \figref{fig:overview}. 

The rest of the paper is organized as follows. First, we discuss how to define the centroid uncertainty in \secref{sec:measurement}. Based on this uncertainty, we introduce a precise circular pattern detector that guarantees the connectivity of boundary points in \secref{sec:detection}. In \secref{sec:optimization}, the measurement uncertainty is combined with our unbiased conic projection model~\cite{CVPR-2024-song}, establishing a robust loss function for optimization. At last, we navigate what the good camera calibration is and how to perform it in \secref{sec:cal_uncertainty}. The key contributions are summarized as follows.
\begin{itemize}
    \item We are the first to define the centroid uncertainty of an arbitrary two-dimensional shape in the image. The derivation based on probabilistic modeling using the energy function provides mathematical thoroughness.
    \item We propose an uncertainty-aware framework of camera calibration using the circular pattern. The uncertainty-based pattern detection and robust optimization process, combined with our unbiased estimator, significantly enhances the overall accuracy and robustness of results.
    \item We provide guidelines for achieving reliable camera calibration, marking a departure from experience‐based approaches and enabling non‐experts to obtain consistent and accurate results.
\end{itemize}

%% file: sec/2_related_work.tex
\section{Related Works}
\label{sec:related_work}
\subsection{Target-based camera calibration}
The camera calibration is a process of determining the intrinsic parameters, which parameterize the camera model that describes the relationship between rays entering the camera and the corresponding pixel positions in the image. 
Initially, this task required special apparatus and measuring the exact position of the targets in 3D space \cite{MIT-1993-faugeras,RAL-1987-tsai}, which are laborious for non-experts.
To mitigate this issue, \citet{TPAMI-2000-zhang},  \citet{CVPR-1999-sturm} introduced a calibration method utilizing a planar target with some repeated pattern printed on it. They eliminated the need to pre-measure the target's position, leveraging the planar characteristics that points on a plane satisfy a homography relationship with the points in the image. The points utilized in the calibration are called control points, and the pattern is used to extract the position of control points in the image.
The most widely used pattern is the checkerboard, whose control points are the corners of black-and-white squares. However, the locations of control points in images exhibit pixel level errors due to the discretized image pixels. 

To overcome the limitation of points as zero-dimensional features, the use of high-level features such as lines and conics has gained momentum. \citet{MVA-2001-devernay} leveraged the line feature, which is always straight in undistorted images. Although this work did not explicitly estimate the intrinsic parameters, recent work~\cite{IP-2021-chuang} introduced the closed-form solution of the intrinsic parameters using undistorted images. The line features are more robust than points; however, their 3D representations often assume infinite lines and have degeneracy in the parallel direction. Therefore, their usage is limited to undistorting the image or extrinsic calibration~\cite{RAL-2019-jeong, ICRA-2024-shin}.

The two-dimensional feature, conic, benefits from its complete matrix representation. \citet{TPAMI-2000-heikkila} introduced a camera calibration pipeline using circular patterns, whose control points are the centroids of projected circles. These control points are robust to image noise and provide sub-pixel level accuracy~\cite{ICPR-1996-heikkila}. 
However, it shows poor calibration performance due to the erroneous projection model of the centroid. Specifically, the circle centers in the target are not projected to the centroid of the projected circle in the image. To fully exploit the conic feature, mathematics for projecting the entire 2D shape is required, instead of the point projection model. Several works~\cite{TPAMI-2000-heikkila,TPAMI-2005-kim, ECCV-2004-chen} addressed this problem by adopting a conic-based transformation. These methods achieved the unbiased projection model under linear transformations, such as perspective transformation. However, the bias introduced by lens distortion is unresolved since conic characteristics are not preserved under this nonlinear transformation. To resolve this issue, \citet{TPAMI-2006-kannala} introduced a generalized concept of the unbiased projection model for the circular pattern but could not derive an analytic solution for the given integral equation.

Since the error arising from the biased projection of conic is more dominant than the error originating from inaccurate feature points on the checkerboard~\cite{PAL-2007-mallon}, previous studies have focused on improving the performance of the checkerboard-based method~\cite{ICCV-2009-datta, IM-2021-vidas}. However, \citet{CVPR-2024-song} revisits the inherent advantages of conic features and brings a significant breakthrough by proposing the unbiased conic projection model under distortion. This work proves that the circular pattern is superior to the checkerboard in both robustness and accuracy.  
Nevertheless, other crucial parts for circle-based camera calibration, such as the detection method and robust optimization, have not been discussed as much as the checkerboard-based methods.

\subsection{Uncertainty in camera calibration}
In practice, image pixels do not adhere to the ideal camera model.
Hence, several works~\cite{RAL-1987-tsai, ICCV-2001-grossberg, TPAMI-2016-ramalingam} tried to increase the model complexity by suggesting general projection models, such as B-spline interpolation with nearest points~\cite{CVPR-2020-schops} or pixel-wise focal length~\cite{CVPR-2022-pan}. 
However, due to the trade-off between model complexity and overfitting\footnote{In other words, it is called the bias-variance trade-off.}, these approaches are susceptible to outliers and require a large amount of data. Moreover, existing applications still rely on the simplicity and linearity of fundamental camera models.

To address the inherent error of camera models and the potential measurement inaccuracy while maintaining the conventional approach, recent studies~\cite{IROS-2013-richardson, ICCV-2019-Peng, IJCV-2022-hagemann, TRO-2024-Ulrich, ICRA-2013-Ozog} introduced a probabilistic approach to the calibration. They model the intrinsic parameters as a random variable and provide the covariance matrix of the parameters, rather than solely estimating the mean values. This calibration uncertainty plays a key role in evaluating the calibration results~\cite{IJCV-2022-hagemann} or suggesting the next-best-view for calibration by minimizing the uncertainty of the estimated camera parameters~\cite{IROS-2013-richardson, ICCV-2019-Peng}. 

The mainstay of this approach is the appropriate modeling of measurement uncertainty (\ie the position uncertainty of observed control points). However, the literature assumes the standard normal distribution~\cite{IROS-2013-richardson,  IJCV-2022-hagemann, ICRA-2013-Ozog} or relies on empirical assumptions without theoretical derivation~\cite{ICCV-2019-Peng}. Furthermore, these works only focus on point uncertainty in the checkerboard-based method, and the centroid uncertainty in the circular pattern has not been discussed. Although the centroid is a point, its derivation based on a two-dimensional shape necessitates a fundamentally different approach to define its uncertainty.

%% file: sec/3_Shape_unc.tex
\section{Measurement}
Camera calibration begins with obtaining measurements from images, which have associated uncertainty. Incorporating the measurement uncertainty appropriately at each calibration stage is the key to performing the optimal calibration from the given data. Hence, we will first discuss the measurement and its uncertainty before proceeding with the calibration process. In this section, a vector $\bs{p} \in \Re^n$ and a matrix $\bs{Q} \in \Re^{n\times m}$ are denoted by lowercase and uppercase in bold.
\subsection{Measurement model: centroid of $n$-sided polygon}
\label{sec:measurement}
\begin{figure}[!t]
    \centering
    \includegraphics[trim=22 40 30 35, clip,width=1.0\columnwidth]{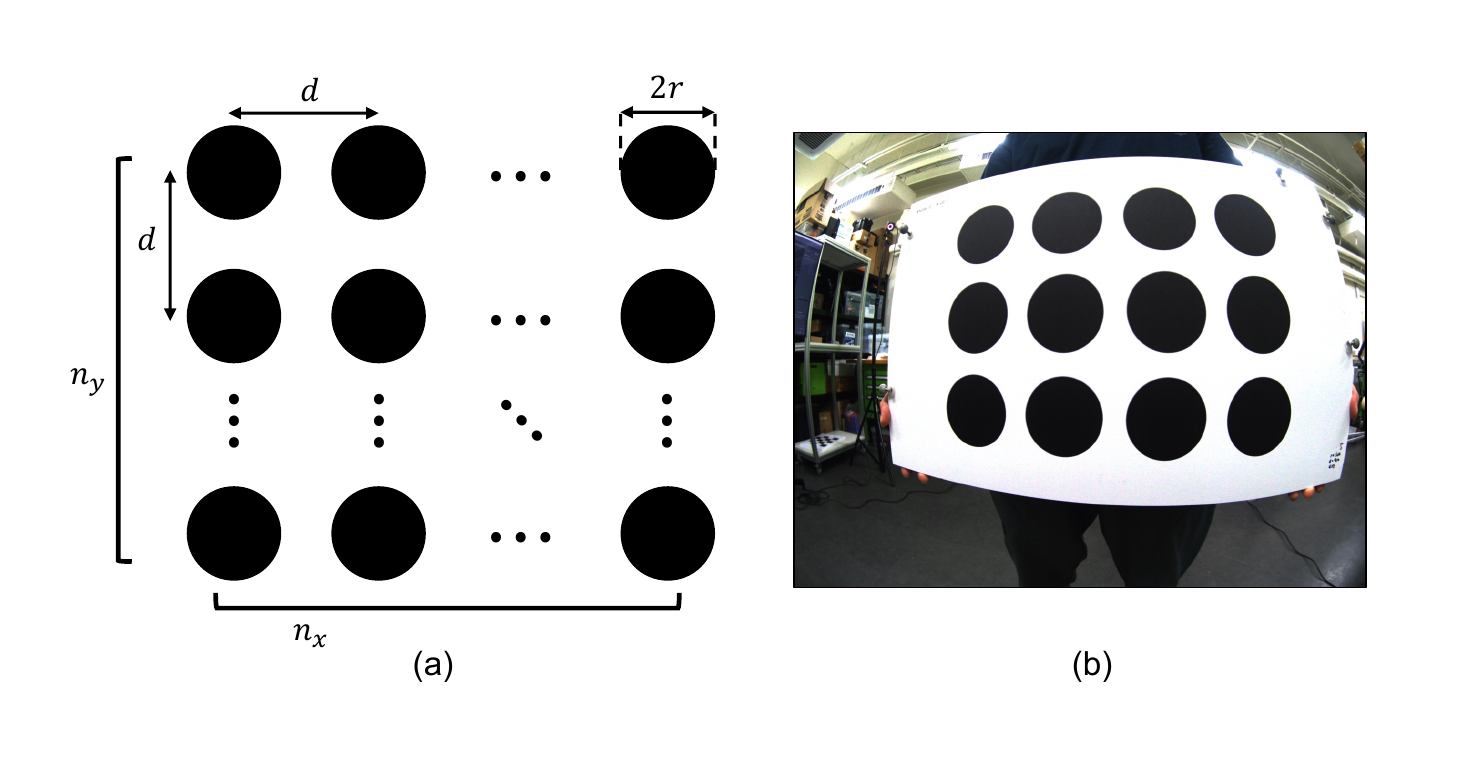}
    \caption{\textbf{Target design}. (a) is a circular pattern printed on a target, and (b) is a sample target image obtained for calibration.}
    \label{fig:target}
\end{figure}

Our calibration target is a grid of circles as \figref{fig:target} (a). The ways to distinguish the circles from the background vary with sensor types. Black circles with white backgrounds are sufficient for most cameras, but sensors like \ac{TIR} cameras, which operate outside the visible spectrum, distinguish circles based on other properties such as emissivity rather than color~\cite{Measure-2017-usamentiaga}. For the sake of consistency, we will assume black circles on a white planar target.

The circular pattern is projected into the image as black blobs as \figref{fig:target} (b), and our measurement model is the centroid of the blobs. Due to the quantized grid structure of image pixels, an arbitrary shape is represented by an $n$-sided polygon in the image space. The centroid of a polygon could be obtained by averaging all inside pixel positions. However, only the boundary points define the entire shape, and interior points do not provide information. In order to define the uncertainty of the centroid, we express the centroid with only boundary points, the determining factor. Based on the Green theorem~\cite{PR-1996-yang}, the centroid of a polygon is expressed as follows.
\begin{subequations}
\begin{align}
    m^{00} &= \frac{1}{2}\sum_{i=1}^n u_{i+1}v_i-u_iv_{i+1},\\
    m^{10} &= \frac{1}{6}\sum_{i=1}^n (u_{i+1}v_i-u_iv_{i+1})(u_{i+1}+u_i),\\
    m^{01} &= \frac{1}{6}\sum_{i=1}^n (u_{i+1}v_i-u_iv_{i+1})(v_{i+1}+v_i),\\
    \bs{p} &= \begin{bmatrix}
        \bar{u}\\
        \bar{v}
    \end{bmatrix} = \frac{1}{m^{00}}\begin{bmatrix}
        m^{10}\\
        m^{01}
    \end{bmatrix}.
\end{align}
\end{subequations}
$\bs{p}$ is the centroid and $\bs{x}_i = (u_i, v_i)$ is a boundary point of the polygon. The index $i$ should be assigned in counter-clockwise (clockwise order in image space).

\begin{figure*}[!t]
    \centering
    \includegraphics[trim=20 10 50 10, clip,width=0.9\textwidth]{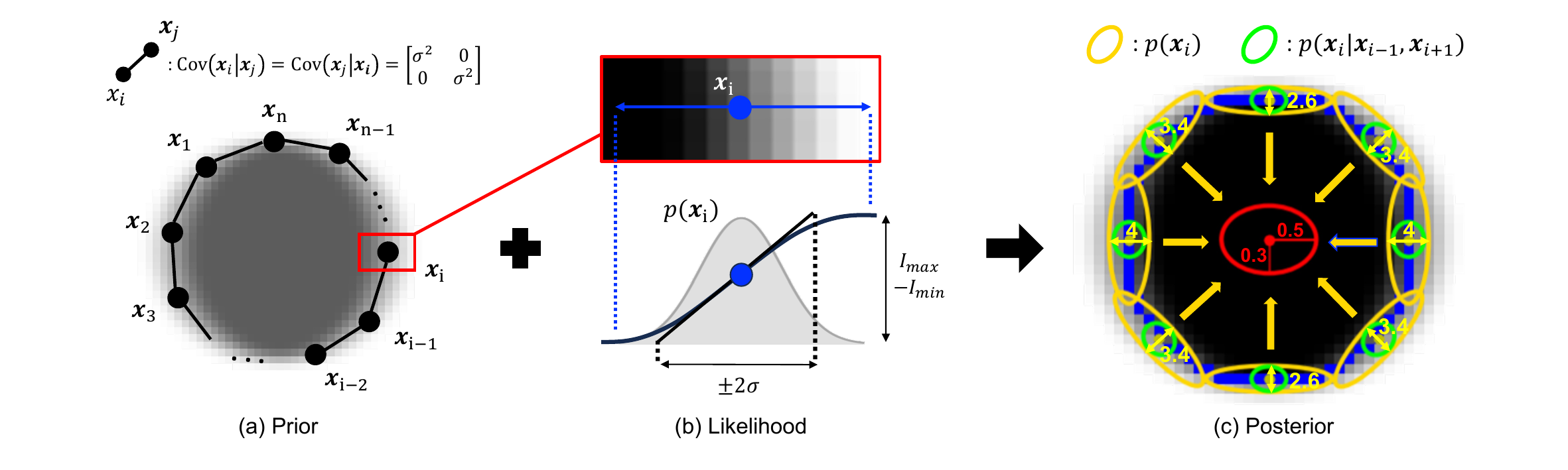}
    \caption{ (a) We model the connectivity of the $n$-sided polygon as a Markov random field. Indeed, every $x_i$ follows a zero-mean Gaussian distribution by subtracting the mean values. But we have displayed it at the mean position for visualization convenience.
    (b) Image gradient limits the freedom of boundary points in its direction. This information is transmitted to the prior distribution. (c) The uncertainty of boundary points is propagated to the uncertainty of the centroid. 
    While the minor axis length of the covariance of boundary points averages around 3, the major axis length of the centroid’s covariance is less than 1. Note that the distributions of boundary points span the ambiguous region (blurry region in the boundary).}
    \label{fig:measurement_uncertainty}
\end{figure*}

\subsection{Shape uncertainty}
The boundary points could be obtained by gradient-based contour detection methods~\cite{TPAMI-1986-canny, TPAMI-1986-Torre} such as Canny edge detector. These methods seek points whose gradients are local maximum; however, it does not imply the points are ground truth. The position of a boundary point is a random variable, and a high gradient implies high probability. We denote the distribution of boundary points $\Tilde{\bs{X}}=[\Tilde{\bs{x}}_1, \Tilde{\bs{x}}_2, \cdots, \Tilde{\bs{x}}_n]^{\top}$ as
\begin{equation}
    P(\Tilde{\bs{X}}| I) \sim \mathcal{N}(\bar{\bs{X}}, \bs{\Sigma}),
\end{equation}
which is a multi-variable Gaussian distribution whose mean vector $\bar{\bs{X}}$ is a detection result. $I$ indicates the input image.

There exist several methods defining the uncertainty of a point \cite{ICCV-2019-Peng}. However, applying a similar approach to shape boundaries is not valid, as the boundary points are not independent but highly connected. 
For instance, when pixels $i$ and $j$ are consecutive boundary points, the difference between their $x$ (also $y$) coordinates must be at most one. To reflect this connectivity, we separate the posterior distribution $P(\Tilde{\bs{X}}| I)$ into two terms as
\begin{align}
    P(\Tilde{\bs{X}}| I)&\varpropto P(\Tilde{\bs{X}})P(I|\Tilde{\bs{X}}),
\end{align}
where $P(\Tilde{\bs{X}})$ is the prior distribution of boundary points and $P(I|\Tilde{\bs{X}})$ is the likelihood. We first establish the covariance of $P(\Tilde{\bs{X}})$ and update the covariance using the likelihood term based on the Information filter.


\subsection{Prior distribution of shape}
For the convenience of the mathematical derivation, we introduce
$\bs{X} = \Tilde{\bs{X}} - \bar{\bs{X}}$ to transform the original distribution into a zero-mean distribution. The connectivity condition can be expressed probabilistically as follows.
\begin{equation}
    \label{eq:pxixj}
    p(\bs{x}_i|\bs{x}_j) = p(\bs{x}_j|\bs{x}_{i}) \sim \mathcal{N}\left(\bs{0},\begin{bmatrix}
        \sigma^2 &0\\
        0 & \sigma^2
    \end{bmatrix}\right).
\end{equation}
A hyperparameter $\sigma$ controls the deformation flexibility of the detected contour. The smaller $\sigma$ means the stronger connection between boundary points, causing the contour to remain fixed in its initial structure. Conversely, as $\sigma$  increases, boundary points are allowed to move more freely without necessarily being tightly connected. Consequently, the distribution of boundary points is expressed as an unidirectional Markov random field as \figref{fig:measurement_uncertainty} (a).

Since the prior distributions of $x$-coordinates and $y$-coordinates are independent and identical, it is sufficient to consider just one variable ($x$-coordinates) as
\begin{equation}
    \label{eq:puiuj}
    p(u_i|u_j) = p(u_j|u_{i}) \sim \mathcal{N}(0,\sigma^2).
\end{equation}
Without information from the image, the contour should be free to undergo rigid body motion\footnote{Indeed, the rigid body motion is bounded by the image size. However, the likelihood term $P(I|\Tilde{\bs{X}})$ restricts the motion at the pixel level, rendering this condition meaningless.} (\ie translation and rotation) as 
\begin{equation}
    \label{eq:pxi}
    p(u_i) \sim \mathcal{N}(0,\infty) \text{ for all } i \in \{1,2, \cdots, n\}.
\end{equation}
The infinity term in \eqoref{eq:pxi} implies that defining $\bs{\Sigma}$ is infeasible; however, defining the information matrix, the inverse of covariance, is possible. The information form takes advantage of reflecting conditional probability and infinite covariance.
\begin{lemma}
    \label{lemma:information}
    When the information form of the joint normal distrtion $P(\bs{\alpha}, \bs{\beta})$ is $\mathcal{N}^{-1}\left(\bs{0},\begin{bmatrix}
        \bs{\Omega}_{\alpha\alpha} & \bs{\Omega}_{\alpha\beta}\\
        \bs{\Omega}_{\beta\alpha} & \bs{\Omega}_{\beta\beta}
    \end{bmatrix}\right)$, it satisfies followings.
    
    1) Conditioning:
    \begin{equation}
         P(\bs{\alpha}|\bs{\beta}) \sim \mathcal{N}^{-1}(\bs{0}, \bs{\Omega}_{\alpha\alpha})
    \end{equation}

    2) Marginalization:
    \begin{equation}
        P(\bs{\alpha}) \sim \mathcal{N}^{-1}(\bs{0}, \bs{\Omega}_{\alpha\alpha}-\bs{\Omega}_{\alpha\beta}\bs{\Omega}_{\beta\beta}^{-1}\bs{\Omega}_{\beta\alpha} )
    \end{equation}
\end{lemma}
\noindent Using \lemref{lemma:information}, the information matrix of case $n=2$ is obtained as 
\begin{equation}
    \label{eq:n2info}
    \bs{\Omega}=\frac{1}{\sigma^2}\begin{bmatrix}
        1 & -1\\
        -1 & 1
    \end{bmatrix},
\end{equation}
which satisfies \eqtref{eq:puiuj}{eq:pxi}. 

Expanding \eqoref{eq:n2info} to the general case ($2\rightarrow n$) is not straightforward. We resolve this problem by introducing an energy function. According to Hammersley–Clifford theorem, the probability distribution of the unidirectional Markov random field is also expressed as \
\begin{align}
    \label{eq:energy_form}
    P(\bs{X}) \sim \mathcal{N}(\bs{\mu}, \bs{\Omega}^{-1})=\frac{1}{Z}\text{exp}\left[-\sum_{c_i\in C}f_i(c_i)\right],
\end{align}
where $c_i$ is a clique\footnote{Clique means an undirected fully connected subgraph} and $f_i$ is an energy function. $Z$ is normalization factor and $C$ is set of all possible cliques.
\begin{lemma}
    \label{lemma:info_full}
    If $\bs{X}=[u_1, \cdots, u_n]^{\top}$ is a fully connected Markov random field satisfying $p(u_i|u_j) \sim \mathcal{N}(0,\sigma^2)$ for all $(i,j)$, its information matrix $\bs{\Omega}$ is formed as
    \begin{equation}
        \bs{\Omega} = \frac{2}{n\sigma^2}\begin{bmatrix}
        n-1 & -1 & \cdots & -1\\
    -1 & n-1 &  \cdots & -1\\
    \multicolumn{4}{c}{\vdots}\\
    -1 & -1 & \cdots & n-1\\
    \end{bmatrix}.
    \end{equation}
\end{lemma}
\begin{proof}
    See \bl{Appendix A}.
\end{proof}

\begin{thm}
    \label{thm:energy_full}
    If $\bs{X}=[u_1, \cdots, u_n]^{\top}$ is a fully connected Markov random field satisfying $p(u_i|u_j) \sim \mathcal{N}(0,\sigma^2)$ for all $(i,j)$, its energy function is formed as
    \begin{equation}
        f(\bs{X}) =\frac{1}{n\sigma^2} \sum_{(\forall u_i, u_j\in \bs{X})} (u_i-u_j)^2.
    \end{equation}
\end{thm}

\begin{proof}
\footnotesize
\begin{align}
    P(\bs{X}) 
    &= \frac{1}{Z}\text{exp}\left[-\frac{1}{2}\bs{X}^{\top}\bs{\Omega}\bs{X}\right] \hspace{5mm}\text{(by \lemref{lemma:info_full})}\\
    & = \frac{1}{Z} \text{exp}\left[-\frac{1}{n\sigma^2}\bs{X}^{\top}\begin{bmatrix}
        n-1 & -1 & \cdots & -1\\
        -1 & n-1 &  \cdots & -1\\
        \multicolumn{4}{c}{\vdots}\\
        -1 & -1 & \cdots & n-1\\
    \end{bmatrix}\bs{X}\right]\\
    & = \frac{1}{Z} \text{exp}\left[-\frac{1}{n\sigma^2}\sum_{(\forall u_i, u_j\in \bs{X})} (u_i-u_j)^2\right]\\
    &= \frac{1}{Z}\text{exp}\left[-\sum_{c_i\in C}f_i(c_i)\right]
    \end{align} 
\normalsize
\end{proof}

In our problem, node $u_i$ is only connected to $u_{i-1}$ and $u_{i+1}$; therefore, the maximum length of cliques is two (\eg a subgraph comprising $u_i$ and $u_{i+1}$). \thmref{thm:energy_full} and \eqoref{eq:energy_form} imply that the probability distribution of this Marchov random field can be written as the sum of the squared differences between the $x$-coordinates of all consecutive points.
Consequently, the $n\times n$ information matrix of the prior distribution is derived as 

\vspace{-10pt}
\small
\begin{equation}
    \label{eq:prior}
    \bs{\Omega}_{prior\_x}= \frac{1}{z\sigma^2}\begin{bmatrix}
        2 & -1 &0&0& \cdots & 0&-1\\
        -1& 2 &  -1&0& \cdots &0& 0\\
        0 & -1 & 2 &-1& \cdots &0& 0 \\
        0 & 0 & -1 &2& \cdots &0& 0 \\
        \multicolumn{7}{c}{\vdots}\\
        0 & 0 &0&0& \cdots & 2&-1\\
        -1 & 0 &0&0& \cdots & -1&2\\
    \end{bmatrix},
\end{equation} 
\normalsize
where $z$ is normalization factor determined by \eqoref{eq:pxixj}. This value is a function of $n$, converging to $3-\sqrt{2}$ when $n$ goes to infinity (see \bl{Appendix B}). In most cases, the total number of boundary pixels is sufficiently large; hence, we fixed this value as a constant.

We have only considered $x$-coordinates. By extending \eqoref{eq:prior} to both the $x$ and $y$ coordinates using the previously mentioned i.i.d. condition, the final prior information matrix is expressed as

\vspace{-10pt}
\small
\begin{equation}
    \label{eq:prior_total}
    \bs{\Omega}_{prior}= \frac{1}{z\sigma^2}\begin{bmatrix}
        2 & 0 &-1&0& \cdots & -1&0\\
        0& 2 &  0&-1& \cdots &0& -1\\
        -1 & 0 & 2 &0& \cdots &0& 0 \\
        0 & -1 & 0 &2& \cdots &0& 0 \\
        \multicolumn{7}{c}{\vdots}\\
        -1 & 0 &0&0& \cdots & 2&0\\
        0 & -1 &0&0& \cdots & 0&2\\
    \end{bmatrix}
\end{equation} 
\normalsize
where $\bs{\Omega}_{prior}$ is a $2n \times 2n$ symmetric matrix.
Note that the covariance $\bs{\Sigma} = \bs{\Omega}^{-1}$ still diverges to infinity due to zero determinant of $\bs{\Omega}$. It implies that the set of boundary points is free from rigid body motion, being limited only to structural deformation.



\subsection{Update information using image gradient}
\label{sec:grad_info}
The image gradient supplements the missing information to $\bs{\Omega}_{prior}$ such as constraint of rigid body motion. With linear approximation as \figref{fig:measurement_uncertainty} (b), the standard deviation $\sigma$ of the boundary point $\bs{x}$ in the image gradient direction is inversely proportional to the gradient norm as
\begin{equation}
    \sigma = \frac{I_{max}-I_{min}}{4||\nabla I(\bs{x)}||_2}.
    \label{eq:point_cov}
\end{equation}
$I_{max}-I_{min}$ implies the intensity difference between the brightest and darkest values. The ideal value is 255, which is rare in the real world when illumination variation is frequent. For robustness, we obtain this value by searching a local window centered around the boundary point.

Since the gradient vector only provides us the uncertainty of single direction, the uncertainty of the orthogonal direction diverges to infinity as the zero gradient. Therefore, we express the distribution using an information matrix as,
\begin{subequations}\label{eq:info_i}
\begin{align}
    \bs{g}_i &= \nabla I(u_i,v_i),\\
    \hat{\bs{g}}_i &= \bs{g}_i / ||\bs{g}_i||_2,\\
    \bs{V} &= \begin{bmatrix}
        \hat{\bs{g}}_i & \hat{\bs{g}}_i^{\perp}
    \end{bmatrix},\\
    \bs{\Omega}_i &= \left(\frac{4||\bs{g}_i||_2}{I_{max}-I_{min}}\right)^2\bs{V} \begin{bmatrix}
        1 & 0\\
        0 & 0
    \end{bmatrix}\bs{V}^{\top}.
\end{align}
\end{subequations}


The remaining step is to update the prior information matrix $\bs{\Omega}_{prior}$ using $\bs{\Omega}_{i}$. Based on the Information filter, the posterior information matrix is derived as 
\begin{equation}
    \bs{\Omega}_{posterior} = \bs{\Omega}_{prior} + \begin{bmatrix}
        \bs{\Omega}_1 & 0 & \cdots & 0\\
        0 & \bs{\Omega}_2 & \cdots & 0\\
        \multicolumn{4}{c}{\vdots}\\
        0& 0 & \cdots & \bs{\Omega}_n
    \end{bmatrix},
\end{equation}
where  $\bs{\Omega}_{prior}$ is $2n \times 2n$ matrix from~\eqoref{eq:prior_total}. Note that though the information matrix of the prior and likelihood term is not invertible, the summation of them $\bs{\Omega}_{posterior}$ is invertible. For the sake of simplicity, we will denote 
$\bs{\Omega}_{posterior}$ as $\bs{\Omega}$ from this point onward.

\subsection{Centroid uncertainty}
\label{sec:centroid_unc}
The next step is to propagate the covariance of boundary points to the centroid $\bs{p}$. As a first-order approximation~\cite{ICCV-2019-Peng, IJCV-2022-hagemann}, the covariance of the centroid $\bs{\Sigma}_{\bs{p}}$ is calculated as follows.

\begin{align}
    \bs{J} &= \begin{bmatrix}
        \frac{\partial \bar{u}}{\partial u_1} & \frac{\partial \bar{u}}{\partial v_1} &
        \frac{\partial \bar{u}}{\partial u_2} & \frac{\partial \bar{u}}{\partial v_2} &
        ... &
        \frac{\partial \bar{u}}{\partial u_n} & \frac{\partial \bar{u}}{\partial v_n}\\
        \frac{\partial \bar{v}}{\partial u_1} & \frac{\partial \bar{v}}{\partial v_1} &
        \frac{\partial \bar{v}}{\partial u_2} & \frac{\partial \bar{v}}{\partial v_2} &
        ... &
        \frac{\partial \bar{v}}{\partial u_n} & \frac{\partial \bar{v}}{\partial v_n}
    \end{bmatrix},\\
    \label{eq:measurement_uncertainty}
    \bs{\Sigma}_{\bs{p}} &= \bs{J}\bs{\Omega}^{-1}\bs{J}^\top.
\end{align}
The matrix $\bs{J}$ is a Jacobian matrix regarding to centroid $\bs{p}$ and boundary points $(u_i, v_i)\in S$. Each component of $\bs{J}$ is calculated as

\small
\begin{subequations}
\begin{align}
    \Delta u_i^j &\triangleq u_{j}-u_i ,\\
    \Delta v_i^j &\triangleq v_{j}-v_i,\\
    (m^{00})'_i &= \left[ \frac{dM^{00}}{du_i},\frac{dM^{00}}{dv_i}\right], \\
    & = \frac{1}{2}\left[ -\Delta v_{i-1}^{i+1}, \Delta u_{i-1}^{i+1}\right],\\
    (m^{10})'_i &= \frac{1}{6}\begin{bmatrix}
        -u_{i+1}\Delta v_i^{i+1}-2u_i\Delta v_{i-1}^{i+1}-u_{i-1}\Delta v_{i-1}^{i} \\ (u_{i+1}+u_i+u_{i-1})\Delta u_{i-1}^{i+1}
    \end{bmatrix}^\top,\\
    (m^{01})'_i &=\frac{1}{6} \begin{bmatrix}
        -(v_{i+1}+v_i+v_{i-1})\Delta v_{i-1}^{i+1} \\ v_{i+1}\Delta u_i^{i+1}+2v_i\Delta u_{i-1}^{i+1}+v_{i-1}\Delta u_{i-1}^{i}
    \end{bmatrix}^\top,\\
    \label{eq:measurement_jacobian}
    \bs{J}_i &= \begin{bmatrix}
        \frac{\partial \bar{u}}{\partial u_i} & \frac{\partial \bar{u}}{\partial v_i}\\
        \frac{\partial \bar{v}}{\partial u_i} & \frac{\partial \bar{v}}{\partial v_i}
    \end{bmatrix}= \frac{1}{m^{00}}\left( \begin{bmatrix}
        (m^{10})'_i \\ (m^{01})'_i
    \end{bmatrix}  - (m^{00})'_i{\bs{p}} \right).
\end{align}
\end{subequations}

\normalsize
\input{sec/3-1_algorithm}
The matrix inversion of the $2n\times 2n$ matrix $\bs{\Omega}$ in \eqoref{eq:measurement_uncertainty} involves considerable computational cost. To reduce the overhead, we utilize Cholesky decomposition as
\begin{subequations}
\begin{align}
    \bs{\Omega}&=\bs{L}\bs{L}^{\top}, \\
    \bs{T} &=   \bs{L}^{-1}  \bs{J}^{\top},\\
    \label{eq:centroid_unc}
    \bs{\Sigma}_{\bs{p}} &= \bs{T}^{\top}\bs{T}.
\end{align}  
\end{subequations}
The entire process to obtain the centroid uncertainty is summarized in \algoref{alg:centroid_uncertainty}.

The advantages of centroids as control points are displayed in \figref{fig:measurement_uncertainty} (c). The centroid exhibits sub-pixel level uncertainty while each point has pixel level uncertainty. This corroborates that the measurement of circular patterns is more accurate than point-based patterns, such as a checkerboard.


By analyzing the Jacobian matrix $J$, we can anticipate how the covariance of a boundary point affects the covariance of the centroid. From \eqoref{eq:measurement_jacobian}, the Jacobian component associated with $\bar{u}$ and $i^{th}$ boundary point $(u_i, v_i)$ is rewritten as 

\tiny
\begin{equation}
    (\bar{u})_i' = \frac{1}{M^{00}} \begin{bmatrix}
        -\left(\frac{u_{i+1}+2u_i}{3} - \bar{u} \right)\frac{\Delta v_i^{i+1}}{2} - \left(\frac{2u_{i}+u_{i-1}}{3} - \bar{u} \right)\frac{\Delta v_{i-1}^{i}}{2}\\
        \left( \frac{u_{i+1}+u_{i}+u_{i-1}}{3}-\bar{u} \right) \frac{\Delta u_{i-1}^{i+1}}{2}
    \end{bmatrix}^T.
\end{equation}
\normalsize
Three terms (\ie, $(u_{i+1}+2u_i)/3$, $(2u_{i}+u_{i-1})/3$, $(u_{i+1}+u_{i}+u_{i-1})/3$) are interpolated points between $u_{i-1}\sim u_{i+1}$, and $|\Delta (u, v)_{i}^j|$ is less than $j-i$ since the boundary points are connected with each other. Therefore, the following theorem is established.
\begin{thm}[Jacobian tendency]
    Let $(u_i, v_i)$ is $i^{th}$ counter-clockwise boundary point of a convex polygon $S$, and the gradient of centroid ($\bar{u}, \bar{v}$) with respect to $(u_i, v_i)$ is $[(\bar{u})'_i, (\bar{v})'_i]^\top$. Then, $(\bar{u})'_i$ and $(\bar{v})'_i$ satisfy the followings.
    
    \footnotesize
    \begin{align*}
    (\bar{u})'_i &\simeq\begin{cases}
        0 &\text{if } |u_i-\bar{u}|<< M^{00} \\
        \frac{u_i - \bar{u}}{2M^{00}} \begin{bmatrix}
            -\Delta v_{i-1}^{i+1}\\ \Delta u_{i-1}^{i+1}
        \end{bmatrix}^T \bot \begin{bmatrix}
            \Delta u_{i-1}^{i+1}\\ \Delta v_{i-1}^{i+1}
        \end{bmatrix}^T  \hspace{-5mm}&\text{otherwise} 
    \end{cases}\\
    ~\\
    (\bar{v})'_i &\simeq
    \begin{cases}
        0 &\text{if } |v_i-\bar{v}|<< M^{00} \\
        \frac{v_i - \bar{v}}{2M^{00}} \begin{bmatrix}
        -\Delta v_{i-1}^{i+1}\\ \Delta u_{i-1}^{i+1}\end{bmatrix}^T
        \bot \begin{bmatrix}
            \Delta u_{i-1}^{i+1}\\ \Delta v_{i-1}^{i+1}
        \end{bmatrix}^T  \hspace{-5mm}&\text{otherwise}
    \end{cases}
    \end{align*}
    \label{thm:jacob_tendency}
    \normalsize
\end{thm}
\begin{proof}
    See \bl{Appendix C}.
\end{proof}

\thmref{thm:jacob_tendency} demonstrates that the boundary points far from centroid in $x$-direction determine $x$-coordinate of the centroid ($\bar{u}$) and $y$-direction determine its $y$-coordinate ($\bar{v}$). Moreover, only the covariance components perpendicular to the contour affect the covariance of the centroid. These characteristics smoothly align with human intuition. We also prove that our centroid uncertainty satisfies translation invariance and rotation equivariance. See \bl{Appendix D}.

%% file: sec/3-1_algorithm.tex
\begin{algorithm}[t!]
    \footnotesize
    \caption{Centroid Uncertainty}
    \label{alg:centroid_uncertainty}
    \hspace*{\algorithmicindent} \textbf{Input:} \\
        \hspace*{\algorithmicindent}\hspace{1em} $I$ : Given intensity image\\
	\hspace*{\algorithmicindent}\hspace{1em} $\{(u_1, v_1), (u_2, v_2), ... (u_n, v_n)\}\in S$ : Detected contour points\\
	\hspace*{\algorithmicindent} \textbf{Output:} \\
        \hspace*{\algorithmicindent}\hspace{1em} $\bs{p}, \bs{\Sigma}$ : Mean and covariance of the centroid.
    \begin{algorithmic}[1]
    \State $Gx$ = Image gradient field in $x$ direction. 
    \State $Gy$ = Image gradient field in $y$ direction. 
    \State $(u_p, v_p) = (u_{n-2}, v_{n-2})$
    \State $(u_c, v_c) = (u_{n-1}, v_{n-1})$
    \State $m^{00}, m^{10}, m^{01} = 0$
    \State $J^{0} = zero\_array([1,2*n])$
    \State $J^{1} = zero\_array([2,2*n])$
    \State $\bs{\Omega} = \bs{\Omega}_{prior}$ \hspace{2cm}\text{\% \eqoref{eq:prior}}
    \For{$i = 1:n$}
        \State $u_f, v_f = u_i, v_i$
        \State $dxy = u_f*u_c-u_c*v_f$
        \State $m^{00} \mathrel{+}= dxy$
        \State $m^{10} \mathrel{+}= (u_f+u_c)*dxy$
        \State $m^{01} \mathrel{+}= (v_f+v_c)*dxy$
        \State $pos = 2*(i-1)$
        \State $J^{0}[0, pos:pos+2] = [v_p-v_f, u_f-u_p]$
        \State $j1 = -u_f*(v_f-v_c)-2*u_c*(v_f-v_p)-u_p*(v_c-v_p)$ 
        \State $j2 = (u_f+u_c+u_p)*(u_f-u_p)$
        \State $j3 = -(v_f+v_c+v_p)*(v_f-v_p) $
        \State $j4 = v_f*(u_f-u_c) + 2*v_c*(u_f-u_p)+v_p*(u_c-u_p)$
        \State $J^{1}[0, pos:pos+2] = [j1,j2]$
        \State $J^{1}[1, pos:pos+2] = [j3,j4]$
        \State $maxI, minI = \text{local\_search}(I,u_c,v_c,window\_size)$
        \State $\bs{\Omega}_i = \text{InfoFromGradient}(I,u_c,v_c,maxI,minI)$  \hspace{0.5cm}\% \eqoref{eq:info_i}
        \State $\bs{\Omega}[pos:pos+2, pos:pos+2] \mathrel{+}= \bs{\Omega}_i$        
        \State $u_p, v_p = u_c, v_c$
        \State $u_c, v_c = u_f, v_f$
    \EndFor
    \State $\bs{p} = [m^{10}/m^{00}/3, m^{01}/m^{00}/3]^\top$
    \State $\bs{J} = (J^1/3-P\_bar\cross J^0)/m^{00}$
    \State $\bs{L} = \text{cholesky\_decomposition}(\bs{\Omega})$
    \State $\bs{T} = solve(\bs{L}, \bs{J}^{\top})$
    \State $\bs{\Sigma} = \bs{T}^{\top} \bs{T}$
    \State \textbf{return} $\bs{p}$, $\bs{\Sigma}$
    \end{algorithmic}
\end{algorithm}

%% file: sec/4_method.tex
\section{Uncertainty-aware calibration}
This section describes how to integrate the measurement uncertainty into the entire calibration process as summarized in \figref{fig:overview}. 
We use $\sim$ for a homogeneous vector representation, thereby $\Tilde{\bs{p}} \simeq  \Tilde{\bs{q}}$ means that two vectors are identical up to scale. All notation rules comply with our previous paper \cite{CVPR-2024-song}.

\subsection{Projection model}
We set the target plane to be the same as the $xy$-plane of the world coordinate. The point on the target plane $\bs{p}_w=(x_w,y_w)^\top$ is projected to a point $\bs{p}_n$ in the normalized plane via perspective projection, and then $\bs{p}_n$ is mapped to a point $\bs{p}=(u,v)^\top$ in image plane under distortion and intrinsic matrix. Adopting a pinhole camera model, $\Tilde{\bs{p}}_w$ is projected to the image as
\begin{eqnarray}
  \bs{K} &=& \begin{bmatrix}
        f_x & \eta & c_x\\
        0 & f_y & c_y\\
        0 & 0 & 1
    \end{bmatrix},\\
  \bs{T}_{cw} &=& \begin{bmatrix}
        \bs{R} & \bs{t}\\
        \bs{0}^{\top} & 1
    \end{bmatrix} = \begin{bmatrix}
        \bs{r}_1 & \bs{r}_2 & \bs{r}_3& \bs{t}\\
        0 & 0 & 0 & 1
  \end{bmatrix}, \\
  \Tilde{\bs{p}} &=&\begin{bmatrix}
        u\\v\\1
    \end{bmatrix}
    \simeq \bs{K}\begin{bmatrix}
        \bs{r}_1 & \bs{r}_2 & \bs{r}_3&\bs{t}\\
    \end{bmatrix}\begin{bmatrix}
        x_w\\
        y_w\\
        z_w\\
        1
  \end{bmatrix} \\
   & \simeq & \bs{K} \begin{bmatrix}
        \bs{r}_1 & \bs{r}_2 & \bs{t}\\
    \end{bmatrix}\begin{bmatrix}
        x_w\\
        y_w\\
        1
    \end{bmatrix}\hspace{2mm} (\because z_w=0)\\
        & \simeq & \bs{K} \bs{E} \Tilde{\bs{p}}_w \hspace{3mm}(\Tilde{\bs{p}}_w \triangleq [x_w, y_w, 1]^{\top})\\
    & \simeq & \bs{K}\Tilde{\bs{p}}_n \simeq \bs{H}\Tilde{\bs{p}}_w .
\end{eqnarray}  
\noindent Here, $\bs{K}$ is the intrinsic matrix of the camera, $\bs{E}$ is the extrinsic parameter between the target coordinate and the camera coordinate, and $\bs{H}$ is the homography matrix. The $z_w$ is zero since the $p_w$ is a point located on the $xy$-plane. The above equation fully determines the relation between the target point $\Tilde{\bs{p}}_w$ and the projected point $\Tilde{\bs{p}}_n$ in the normalized plane.
\begin{figure}[!t]
    \centering
    \includegraphics[trim=10 60 30 45, clip,width=0.99\columnwidth]{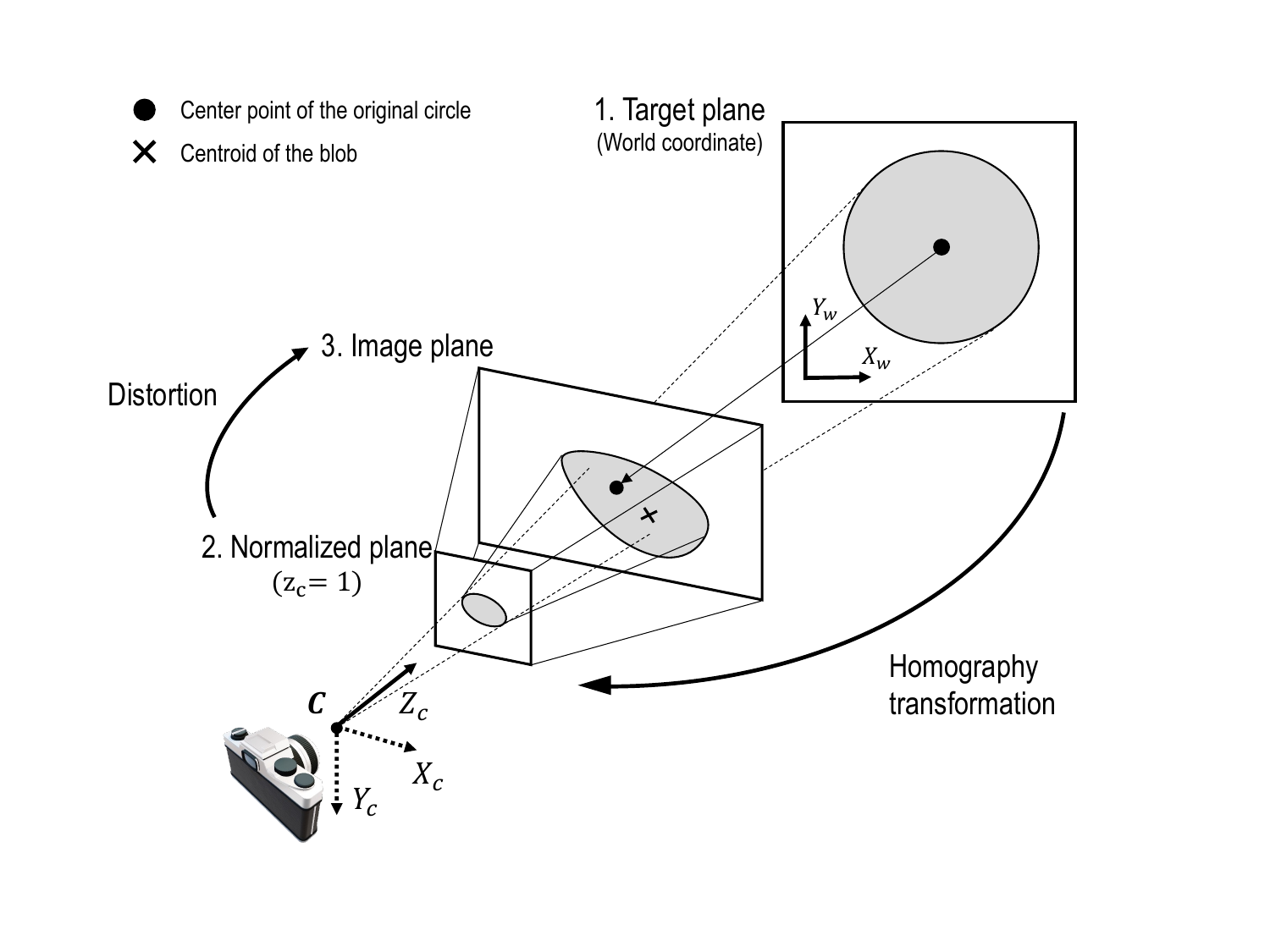}
    \caption{\textbf{The projection model of circular patterns.} 
    A circle on the target plane is projected to the normalized plane of the camera as an ellipse. Then, due to lens distortion, the ellipse is projected to the image as a distorted ellipse. Note that there exists some mismatch between the projected center of the circle on the target plane (circled dot) and the actual centroid of the shape on the image plane (crossed sign). Although the distorted ellipse cannot be analytically described, our estimator \cite{CVPR-2024-song} successfully tracks the true centroid.}
    \label{fig:geometry}
\end{figure}
Unfortunately, this projected point $\Tilde{\bs{p}}_n$ further undergoes a nonlinear transformation (lens distortion) when projected onto the image plane. We only consider radial distortion, which is known to be sufficient in most cameras \cite{CVPR-2019-lopez,CVPR-2022-pan}. The radial distortion is typically modeled with a polynomial function \cite{PE-1996-brown, MNRAS-1919-conrady} as:
\begin{eqnarray}
\label{eq:sn}
s_n &=& x_n^2 + y_n^2,\\
\label{eq:def_dist}
k &=& \sum_{i=0}^{n_d}d_is_n^i \hspace{5mm} (d_0=1),\\
\Tilde{\bs{p}}_d &=& D(\Tilde{\bs{p}}_n)=[x_d, y_d, 1]^\top=[kx_n, ky_n, 1]^\top,\\
\Tilde{\bs{p}}_i &\simeq& \bs{K}\Tilde{\bs{p}}_d = \bs{K}D(\Tilde{\bs{p}}_n) = \bs{K}D(\bs{E}\Tilde{\bs{p}}_w),
\end{eqnarray}
where $d_i$ are the distortion parameters and $D$ is the distortion function. The value $n_d$ is the maximum order of distortion parameters, typically less than three in existing calibration methods~\cite{matlab-2010-bouguet}. 

However, the projection model of circles, conics in 3D space,
is not identical to the point projection model as illustrated in \figref{fig:geometry}. Refer to our paper~\cite{CVPR-2024-song} for the details of unbiased conic projection under lens distortion.
 
\subsection{Uncertainty-based blob detection}
\label{sec:detection}

\begin{figure}[!t]
    \centering
    \includegraphics[trim=20 17 20 23, clip,width=0.9\columnwidth]{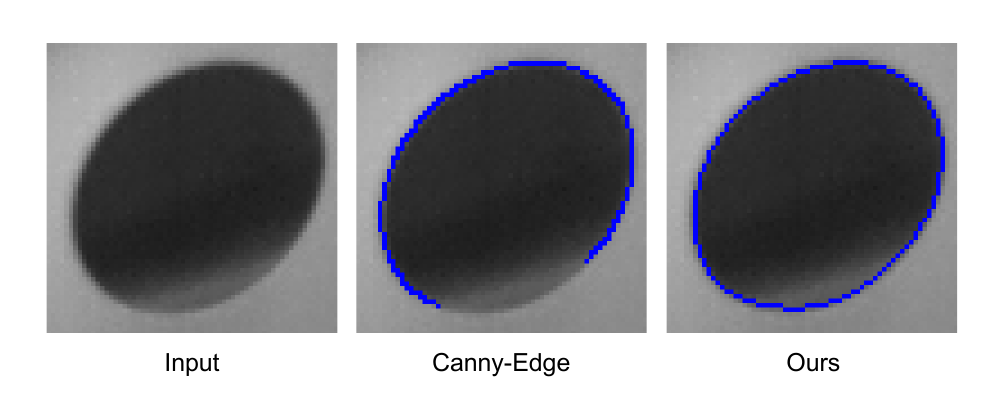}
    \caption{The gradient-based boundary detection method (\eg Canny-Edge detector) does not ensure the connectivity of boundary points. However, our uncertainty-based detector robustly detects the shape boundary even in extreme light conditions.}
    \label{fig:detection}
\end{figure}

For camera calibration, accuracy should be prioritized over computational burden. However, existing detection methods for circular patterns, such as OpenCV blob detection, focus on efficiency and do not scrutinize the boundary points at the pixel level. On the other hand, although existing gradient-based edge detection methods~\cite{TPAMI-1986-canny, AVC-1988-Harris} show high accuracy, they exhibit low detection robustness because image gradients are sensitive to the configuration of light sources. Moreover, they do not ensure the connectivity of boundary points as \figref{fig:detection}. Hence, we have newly developed a calibration-specific circular pattern detector, ensuring both accuracy and robustness.

To guarantee the connectivity of boundary points, we adopt a threshold-based approach. This approach does not directly find the boundary points. Instead, it distinguishes between the interior and exterior regions by applying a threshold to the intensity value, and then identifies the boundary of the two regions. As a result, the connectivity of the boundary points is always guaranteed. Furthermore, this characteristic enables us to enhance the detection results using morphology operations such as closing. The remaining problem is that detection results vary depending on the threshold, and no single optimal value suits every scenario.

We resolve this limitation by selecting the most reliable result using the centroid uncertainty defined in \secref{sec:measurement}. We obtain candidates of measurements as much as possible in various thresholds and then select the blob whose uncertainty is the lowest. The scalar uncertainty used here was derived from the following equation.
\begin{equation}
    \label{eq:uncertainty}
    \epsilon_{\bs{p}} = 2 \sqrt{\text{tr}(\bs{\Sigma}_{\bs{p}})/2} , 
\end{equation}
where $\epsilon_{\bs{p}}$ is the amount of uncertainty of centroid $\bs{p}$, indicating averaged $2\sigma$ range of centroid location.
Instead of the matrix trace, the matrix determinant could be utilized based on information theory (\ie the Shannon entropy of normal distribution). However, the determinant tends to underestimate uncertainty when one eigenvalue is very low, as the determinant is the product of all eigenvalues. The matrix trace is the sum of eigenvalues; therefore, it is more conservative in representing uncertainty.

By combining the threshold-based detector and uncertainty certifier, we successfully develop a robust and accurate circular pattern method. \secref{sec:detection_exp} substantiates that our detection strategy works well, and the results align with human intuition. The entire detection process is encapsulated in \algoref{alg:detection}.

\input{sec/4-1_algorithm}

\subsection{Robust optimization with unbiased estimator}
\label{sec:optimization}
Using the detection results, the initial value of the intrinsic parameters is calculated in closed form using Zhang's method~\cite{TPAMI-2000-zhang}. The initial value is unusable as it does not consider the distortion parameter, requiring refinement through optimization. In the optimization process, we minimize the reprojection error, which is the squared difference between the measurements and the estimated position of the control point using our unbiased estimator \cite{CVPR-2024-song} as
\begin{align}
    (\hat{u}^{jk}, \hat{v}^{jk}) &= \text{UnbiasedEstimator}(\rho, \bs{K}, \bs{E}^j, \bs{Q}^{jk}_w), \\
    \label{eq:loss}
    \mathcal{L} &= \sum_{j=1}^n \sum_{k=1}^m(\bar{{u}}_{jk}-\hat{{u}}_{jk})^2+ (\bar{{v}}_{jk}-\hat{{v}}_{jk})^2, \\
    \bs{K}, \bs{D} &= \argmin_{\bs{K},\bs{D},{\bs{E}^{1,2,...n}}} \mathcal{L},
\end{align}
where the $\bs{p}^{jk} = (\bar{u}^{jk},\bar{v}^{jk})$ is the $k^{th}$ control point in the $j^{th}$ image, and the  $(\hat{u}_i^{jk},\hat{v}_i^{jk})$  is the estimated control point corresponding to $\bs{p}^{jk}$. The circle in target plane corresponding to $\bs{p}^{jk}$ is $\bs{Q_w}^{jk}$. 

However, this standard loss function $\mathcal{L}$ is vulnerable to measurement noise, which results in inaccurate solutions and hinders convergence. In order to enhance the robustness and accuracy of the optimization, we apply measurement uncertainty in the loss function as
\begin{equation}
    \label{eq:robust_loss}
    \mathcal{L}_{robust} = \sum_{i=1}^{nm}\begin{bmatrix}
        \bar{u}_i - \hat{u}_i & \bar{v}_i - \hat{v}_i
    \end{bmatrix}\bs{\Sigma}^{-1}_{\bs{p}_i}\begin{bmatrix}
        \bar{u}_i - \hat{u}_i\\
        \bar{v}_i - \hat{v}_i
    \end{bmatrix}.
\end{equation}
The above least-squares problem can be solved by existing optimization tools such as Ceres Solver~\cite{Ceres_Solver}.

\subsection{Calibration evaluation using uncertainty map}
\label{sec:cal_uncertainty}
The calibration has two types of uncertainty: aleatoric and epistemic uncertainty. First, aleatoric uncertainty demonstrates data noise and model complexity. If measurements are inaccurate or the complexity of the camera model is low, high aleatoric uncertainty appears. The aleatoric uncertainty can be mitigated by employing more advanced detection methods and using higher-order distortion coefficients.
The other uncertainty is epistemic uncertainty, which results from a lack of data. This uncertainty arises when insufficient images are acquired or a degenerated condition occurs between poses. It is reduced when the user acquires appropriate image sets. 

Considering both aleatoric and epistemic uncertainties is crucial for evaluating calibration accuracy~\cite{IJCV-2022-hagemann}.
The reprojection error, the most utilized metric, only represents the aleatoric uncertainty. Moreover, as it is proportional to image resolution, its absolute value is ephemeral.
Hence, we adopt the concept of uncertainty map~\cite{IROS-2013-richardson,ICCV-2019-Peng} for evaluating calibration performance.

Based on uncertainty propagation theory such as Gaussian process~\cite{IJNS-2004-Seeger} and Kalman filter~\cite{BE-1960-Kalman}, the covariance of camera parameters (i.e. focal length, principal points, and distortion parameters) is calculated as

\vspace{-10pt}
\small
\begin{align}
    \bs{\Sigma}_{\bs{K},\bs{D}} & \triangleq
    \begin{bmatrix}
        K_{f_xf_x} &  K_{f_xf_y} & K_{f_xc_x} & ... & K_{f_xd_{n_d}}\\
        K_{f_yf_x} &  K_{f_yf_y} & K_{f_yc_x} & ... & K_{f_yd_{n_d}}\\
        \multicolumn{5}{c}{\vdots}\\
        K_{d_{n_d}f_x} &  K_{d_{n_d}f_y} & K_{d_{n_d}c_x} & ... & K_{d_{n_d}d_{n_d}}\\
    \end{bmatrix},\\
    &= (\bs{J}_c^{\top}\bs{\Sigma}_{\bs{p}}^{-1}\bs{J}_c)^{-1},
\end{align}
\normalsize
where $\bs{J}_c$ denotes the Jacobian matrix with respect to the estimated control points and the camera parameters.
The diagonal term $\bs{\Sigma}_{\bs{K},\bs{D}}$ implies an uncertainty of each camera parameter. 

However, this covariance matrix is not sufficient to evaluate the calibration accuracy. A ray entering through the camera lens is projected as a pixel in the image, and camera parameters merely determine this projection model. Therefore, the essential information is the uncertainty of the pixel position corresponding to the given incidence angle (\ie, elevation and azimuth) of the ray, not the uncertainty of camera parameters. Hence, we prepare $n$ rays and estimate the distribution of each projected point corresponding to each ray as follows.
\begin{align}
    \bs{J}_{proj}[i:i+2,:]&= \begin{bmatrix}
        kx_i & 0\\
        0 & ky_i \\
        1 & 0\\
        0 & 1\\
        ky_i & 0\\
        s(\bar{f_x}x_i+\eta y_i) & s(\bar{f_y}y_i) \\
        s^2(\bar{f_x}x_i+\bar{\eta} y_i) & s^2(\bar{f_y}y_i) \\
        \multicolumn{2}{c}{\vdots} \\
        s^{n_d}(\bar{f_x}x_i+\bar{\eta} y_i) & s^{n_d}(\bar{f_y}y_i)
    \end{bmatrix}^\top\\
    \bs{\Sigma}_{unc} &= \bs{J}_{proj} \bs{\Sigma}_{\bs{K}, \bs{D}} \bs{J}_{proj}^\top
\end{align}
where $(x_{i}, y_i)$ is point in normalized plane corresponding to $\theta_i$ (azimuth) and $\phi_i$ (elevation). $\bs{J}_{proj}$ is the Jacobian matrix with respect to the $(x_{i}, y_i)$ and projected points in the image. The definition of $k$ and $s$ are same as \eqtref{eq:sn}{eq:def_dist}. The covariance $\bs{\Sigma}_{unc}$ are interpolated to represent the uncertainty for each image pixel, which is defined as the uncertainty map. The dimension of this matrix is height$\times$width$\times 3$, where the last channel stands for covariance: $K_{xx}$, $ K_{xy}$, and $ K_{yy}$. 

\begin{figure}[!t]
    \centering
    \includegraphics[trim=62 0 20 0, clip,width=0.99\columnwidth]{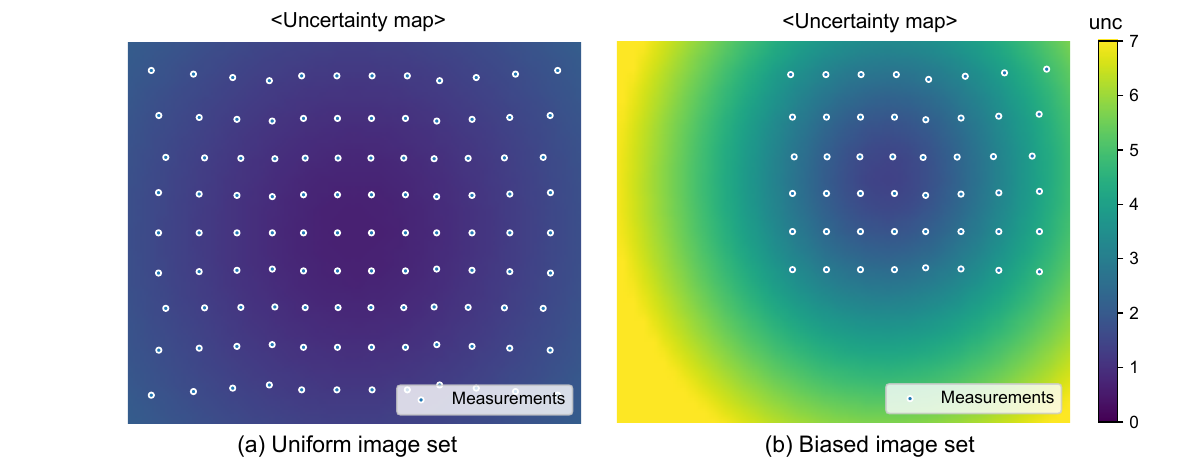}
    \caption{\textbf{Uncertainty map} (a) When we acquire appropriate target images, the uncertainty map shows low uncertainty across the entire image. (b) However, if images are not taken from various poses, the uncertainty distribution is biased in a certain region and shows overall high uncertainty. Unc denotes the scalar uncertainty \eqoref{eq:uncertainty} of each pixel. }
    \label{fig:calibration_uncertainty}
\end{figure}

We illustrate an example of an uncertainty map at \figref{fig:calibration_uncertainty}. The color indicates the scalar uncertainty of each pixel derived from \eqoref{eq:uncertainty}, and white dots denote all aggregated measurements utilized in the calibration.
When the images are not acquired from various poses, the uncertainty is biased towards one side. We will discuss more details about the relationship between poses and calibration uncertainty in \secref{sec:calibration_guide}.


%% file: sec/4-1_algorithm.tex
\begin{algorithm}[t]
    \footnotesize
    \caption{Circular pattern detection}
    \label{alg:detection}
    \hspace*{\algorithmicindent} \textbf{Input:} \\
        \hspace*{\algorithmicindent}\hspace{1em} $I$ : Given intensity image\\
	\hspace*{\algorithmicindent} \textbf{Output:} \\
        \hspace*{\algorithmicindent}\hspace{1em} $\bs{S}=\{S_1, S_2, ... , S_n\}$: Circular shape boundaries in the image.
    \begin{algorithmic}[1]
    \State $\bs{B}= [\,]$
    \State $unc_f = \infty$
    \For{$p :$ parameter candidates }
        \State $I_{thresh} = \text{adaptiveThreshold}(I, p)$
        \State $\bs{C} = \text{findContours}(I_{thresh})$
        \For{$C : \bs{C}$}
            \If{$ellipseTest(C)$}
                \State $\bs{B}.append(C)$
            \EndIf  
        \EndFor
    \EndFor
    \State $\bs{S}=[\,]$
    \For{$C : \bs{B}$}
        \State isnew = true
        \For{$S : \bs{S}$}
            \If{$|centroid(C)-centroid(S)|<threshold$}
                \If{$uncertainty(C)<uncertainty(S)$}
                    \State $S = C$
                \EndIf
                \State isnew = false
                \State break
            \EndIf
        \EndFor
        \If{isnew}
            \State $\bs{S}.append(C)$
        \EndIf
    \EndFor
    \State \textbf{return} $\bs{S}$
    \end{algorithmic}
\end{algorithm}

%% file: sec/5_results.tex
\section{results}
While we have thus far discussed the uncertainty and its incorporation into the calibration process from a theoretical perspective, in this section, we experimentally analyze the practical implications of these components and their impact on calibration performance.
\subsection{Analysis of the measurement uncertainty}
In order to verify that the uncertainty aligns with our intuition and provide further information, we investigate the uncertainty of circular patterns in various scenarios such as Gaussian blur, translation blur (in the $x$-direction), and rotation blur conditions. The motion blurs (\ie translation and rotation blurs) are synthesized based on \cite{SIGGRAPH-2001-brostow}.

\begin{figure}[!t]
    \centering
    \includegraphics[trim=20 17 20 10, clip,width=0.99\columnwidth]{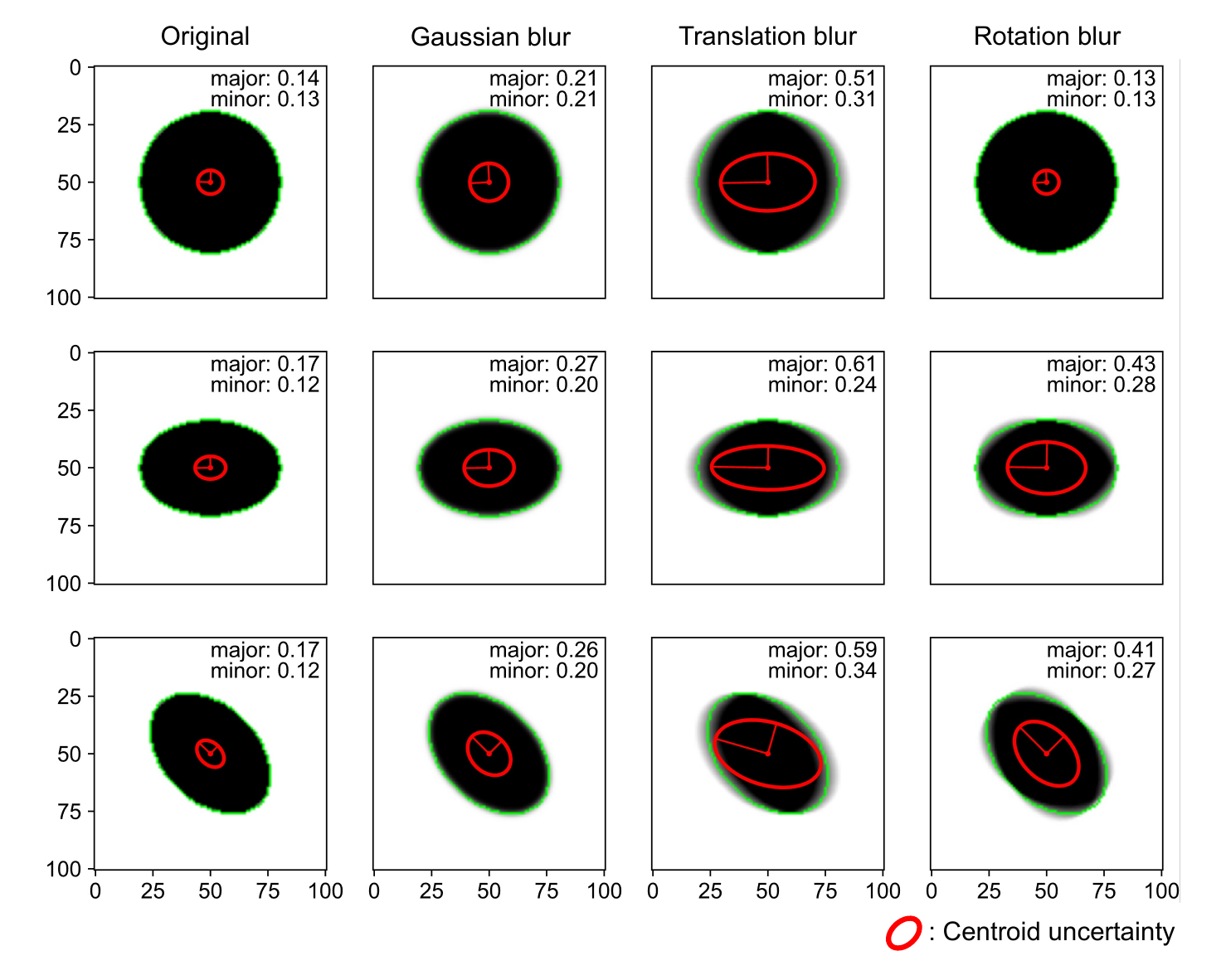}
     \vspace{-5mm}
    \caption{\textbf{Uncertainty tendency.} We investigate how the uncertainty of the centroid behaves in blur situations that may arise during calibration. Since the centroid shows sub-pixel uncertainty, we scale up 40 times for convenience. (Boundary blur: $\sigma=3$, Translation: $\pm 5$ pixel, Rotation: $\pm 20^{\circ}$)}
    \label{fig:circle_unc}
\end{figure}

\figref{fig:circle_unc} illustrates the uncertainty of circles, ellipses, and rotated ellipses, which appear when the circular pattern is projected in the image. The red ellipse represents the $\pm2\sigma$ region of the centroid. As mentioned in \secref{sec:centroid_unc}, only the normal component of the boundary point variance affects the centroid variance; therefore, the final shape of the centroid variance tends to follow the shape of the contour. For instance, ellipses in the second row show low uncertainty in the $y$-direction since the normal vectors of most boundary points of the shape are oriented in the $y$-direction. By comparing the second and third rows of the first two columns, we could confirm that our measurement uncertainty is rotation equivariant. The discrepancy at the second decimal place results from slight changes in the boundary points during the rasterization of a curve-shaped object onto a pixel grid. 

Boundary blur increases overall uncertainty, and motion blur increases the uncertainty in a particular direction more . Despite extreme translation, the uncertainty of the centroid remains under one pixel. 
Interestingly, motion blur in the $x$-direction also increases the uncertainty of $y$ coordinates. This is because the variation of $x$ coordinates alters the area of the local region. For example, if the upper half of an ellipse expands in the $x$-direction, the centroid of the ellipse will shift upward.

\begin{figure}[!t] 
    \centering
    \includegraphics[trim=20 186 20 22,clip, width=0.99\columnwidth]{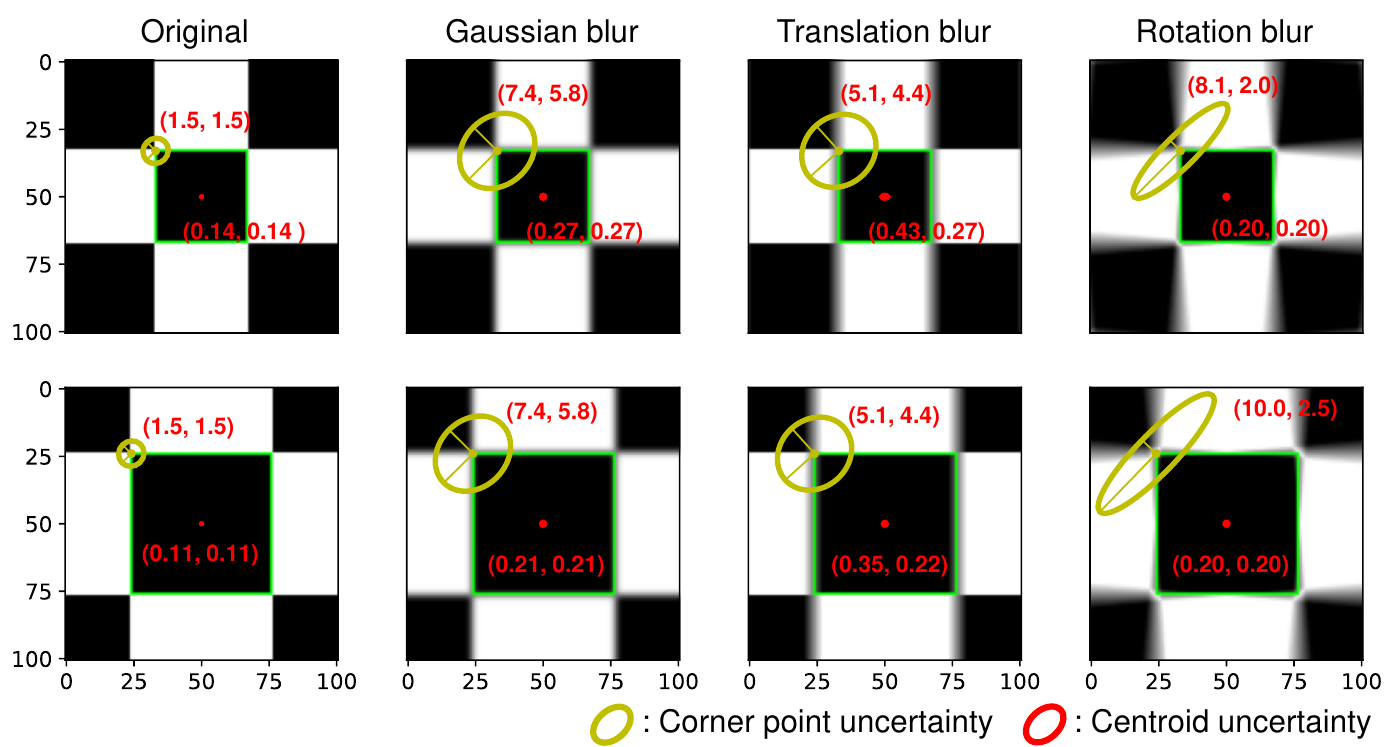}
    \vspace{-5mm}
    \caption{\textbf{Comparison of point and centroid uncertainties.} To clarify the advantage of centroids compared to points, we investigate the uncertainty of square corner points and centroids. The yellow ellipse indicates the uncertainty of the corner point, and the red ellipse indicates the uncertainty of the square centroid. Since point uncertainty is significantly large in the blur conditions, we scale up only three times to visualize both uncertainties simultaneously.
    (Boundary blur: $\sigma=3$, Translation: $\pm 3$ pixel, Rotation: $\pm 5^{\circ}$)
    }
    \label{fig:checkerboard_unc}
\end{figure}

\subsection{Comparison of point and shape uncertainty} 
We demonstrate the potential of the centroids compared to the checkerboard corners in \figref{fig:checkerboard_unc}, illustrating the uncertainty of corner points and the uncertainty of the square centroid simultaneously. The definition of point uncertainty follows the autocorrelation-based method used in \cite{ICCV-2019-Peng}. 
The result clearly substantiates that the centroid is more precise than the corner point and robust to boundary blur effects. 

Another advantage of the centroid is that its uncertainty decreases as the number of boundary points increases. In the first and second rows, while the uncertainty of the corner points remains constant as the size of the rectangle increases, the uncertainty of the centroid decreases. 
In the rotation blur case, point uncertainty becomes bigger when the point is far from the rotation center $(50,50)$. However, the centroid uncertainty remains the same, which demonstrates that the uncertainty of point sets could overcome the innate uncertainty of a single point.


\subsection{Detection based on uncertainty}
\label{sec:detection_exp}
\begin{figure}[!t] 
    \centering
    \includegraphics[trim=120 92 85 67, clip, width=1.0\columnwidth]{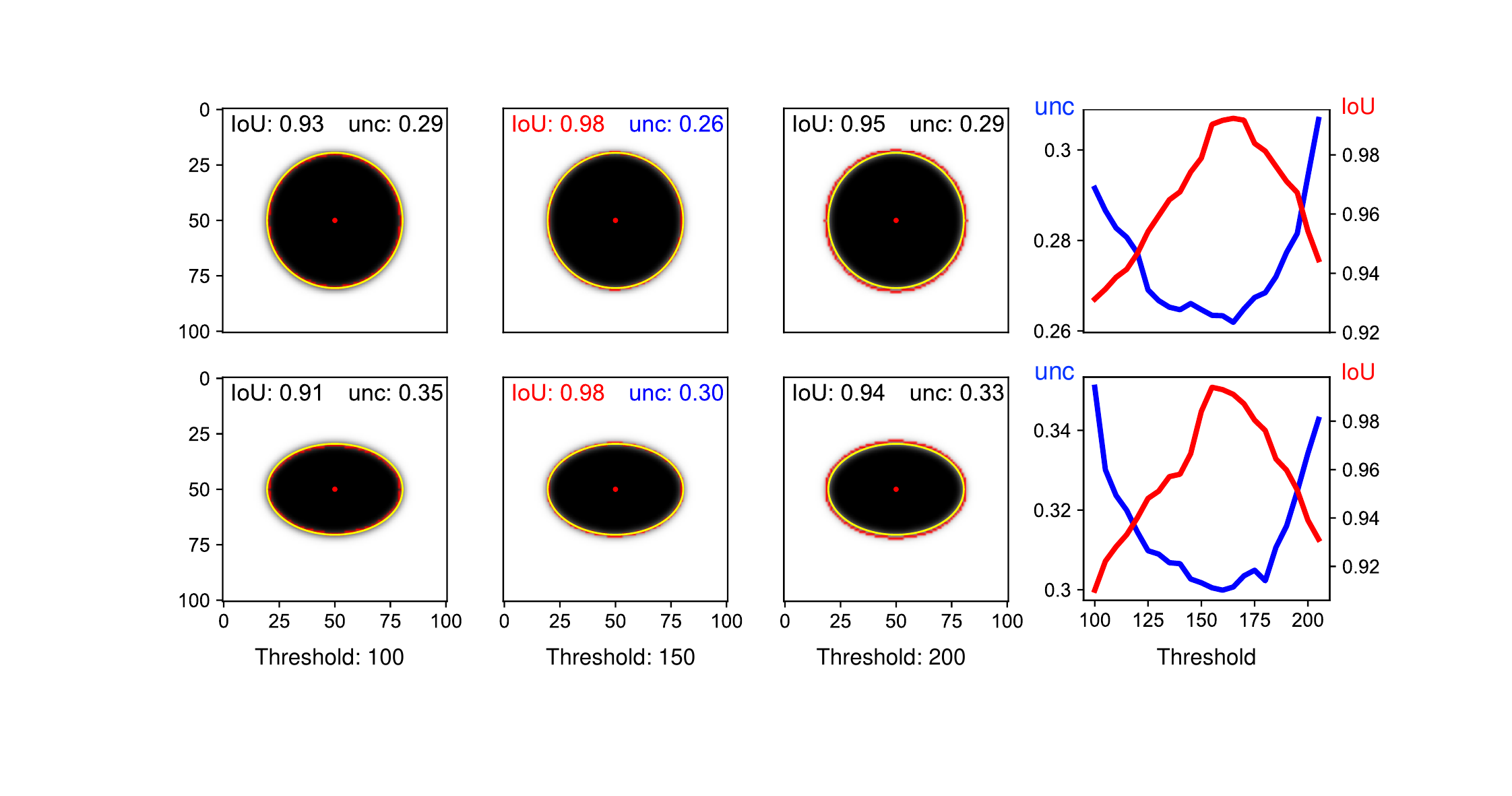}
    \caption{\textbf{Justification of uncertainty-based selection.} We prepare several boundary candidates, increasing the color threshold from 100 to 200. Our uncertainty becomes the lowest when the IoU value between a candidate (red contour) and the ground truth contour (yellow contour) is maximized. This implies we can select the most reliable candidate based on our uncertainty.}
    \label{fig:detection_exp}
\end{figure}

We verify whether our uncertainty-based detector can accurately capture the ground truth contour. We collect contours of given ellipses while increasing thresholds from $100$ to $200$ as \figref{fig:detection_exp}. Since the Gaussian blur is applied three times with $\sigma=3$; the contours highly rely on the threshold values. As a result, low uncertainty corresponds to high \ac{IoU}, which implies that the detected boundary is close to the ground truth. 

The theoretical explanation for this circumstance is that the uncertainty of boundary points decreases when the gradient magnitude increases or the gradient direction aligns more closely with the contour normal. Hence, we could select the boundary that is most aligned with the information from the image gradient field.

\subsection{Ablation study: calibration accuracy}
\begin{figure}[t!]
    \centering
    \includegraphics[trim=10 15 12 20, clip,width=0.9\columnwidth]{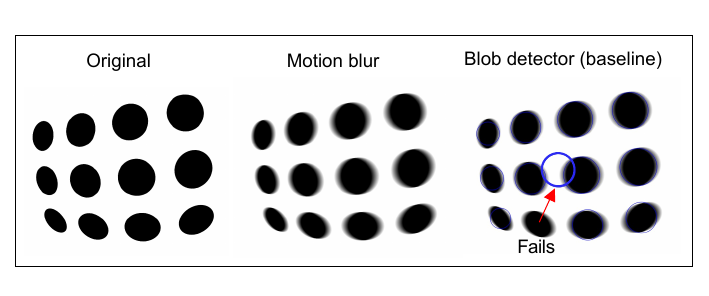}
    \caption{\textbf{Sample images from the synthetic dataset}. The traditional blob detector of \texttt{OpenCV} even fails to detect circles in an ideal synthetic image. }
    \label{fig:cal_results1}
\end{figure}

To verify the efficacy of each module, we conduct ablation studies in two synthetic datasets. The datasets comprise 30 images obtained from random camera poses, and we apply motion blur to only one dataset. The sample image of each dataset is displayed in \figref{fig:cal_results1}.
For calibration, we randomly select six images and repeat 30 times to calculate the mean values and standard deviations of camera parameters (Monte Carlo Method). \tabref{tab:cal_results1} demonstrates the calibration results of each method in the original and motion blur datasets. The baseline is the traditional \texttt{OpenCV} method, whose detector is \texttt{SimpleBlobDetector}. The baseline utilizes the point projection model, ignoring the characteristics of circular patterns. We observe how the calibration results changed as we incrementally added our uncertainty-based detector (D), unbiased-estimator~\cite{CVPR-2024-song} (E), and robust optimization (O) modules to this baseline. \#fails indicates the number of detection failures among the total 30 images.

\begin{table}[t!]
    \centering
    \caption{\textbf{Calibration results of synthetic images.} 
    }
    \label{tab:cal_results1}
    \resizebox{1.0\columnwidth}{!} {
    \renewcommand{\arraystretch}{1.2}
    \begin{tabular}{l|ccccc|c}    
        \hline
        \multicolumn{1}{c|}{Methods} & $f_x$ & $f_y$ &  $c_x$ & $c_y$&  $d_1$  &\#fails\\
        \hline
        \multicolumn{7}{c}{Original}\\
        \hline
        \multicolumn{1}{c|}{GT} & 600 & 600 & 600 & 450 & -0.4&\\
        B (baseline) &   622$\pm $147	&622$\pm $149	&614$\pm $35	&444$\pm $20	&-0.48$\pm $0.30&6\\
        B+D & 603.6$\pm $0.99	&604.6$\pm $0.93	&601.5$\pm $2.16	&450.4$\pm $1.77 & -0.42$\pm $0.004 &0\\
        B+D+E &\textbf{600.2}$\pm$\textbf{0.38}	&\textbf{600.2}$\pm$\textbf{0.35}	&\textbf{600.0}$\pm$0.35	&\textbf{450.0}$\pm$\textbf{0.19}	&\textbf{-0.40}$\pm$\textbf{0.002}&0\\
        B+D+E+O & \textbf{600.2}$\pm$ 0.42 &\textbf{600.2}$\pm$0.39 &\textbf{600.0}$\pm$\textbf{0.34} &\textbf{450.0}$\pm$\textbf{0.19} &\textbf{-0.40}$\pm$\textbf{0.002}&0\\
        \hline
        \multicolumn{7}{c}{Motion blur}\\
        \hline
        B (baseline) & 651$\pm $232	&679$\pm $245	&651$\pm $111	&511$\pm $206	&-0.40$\pm $ 0.93& 6 \\
        B+D & 603.1$\pm $1.35	&603.0$\pm $1.16	&601.2$\pm $1.59	&450.5$\pm $1.78&-0.42$\pm $0.004&0\\
        B+D+E &599.8$\pm$0.69	&599.7$\pm$0.72	&599.7$\pm$1.02	&450.2$\pm$0.38	&\textbf{-0.40}$\pm$\textbf{0.002}&0\\
        B+D+E+O & \textbf{600.1}$\pm$ \textbf{0.49} &\textbf{600.1}$\pm$\textbf{0.48}  &\textbf{599.8}$\pm$\textbf{0.81}  &\textbf{450.1}$\pm$\textbf{0.22} &\textbf{-0.40}$\pm$\textbf{0.002}&0\\
        \hline
        \multicolumn{7}{l}{D: uncertainty-based detector, E: unbiased estimator, O: robust optimization}
    \end{tabular}
    \renewcommand{\arraystretch}{1}
    }
\end{table}

Before adding our detector, the baseline detector often outputs erroneous detection results as \figref{fig:cal_results1}, even in synthetic images. Consequently, the baseline fails to detect in six out of the thirty images, which partially explains why the circular pattern has not been widely used for calibration until now.
Compared with this method, our uncertainty-aware detector significantly improves the detection and calibration accuracy. However, owing to the biased projection, camera parameters converge to erroneous values. 
The combination of our detector and estimator overcomes this issue and shows great performance in the original dataset, while having relatively lower accuracy and high variance in the motion blur dataset. Even for our detector, accurately identifying the ground truth contour under motion blur conditions is challenging. This is why we need to consider measurement uncertainty in the optimization process. Our robust optimizer effectively handles the low-quality measurements, resulting in improved results in the motion blur dataset. Since the images in the original dataset contain only clear measurements, the results are almost identical, regardless of whether uncertainty is considered during the optimization process.

\begin{figure}[t]
    \centering
    \includegraphics[trim=21 25 10 17, clip,width=0.85\columnwidth]{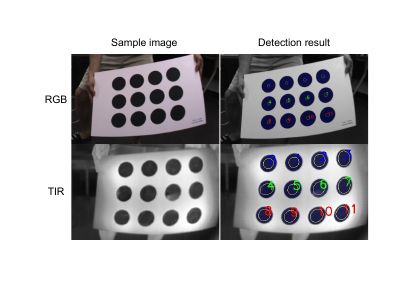}
    \caption{\textbf{Sample images obtained from an RGB camera and a TIR camera.} Our detector not only robustly detects the circular pattern but also outputs the centroid uncertainty, represented as yellow circles. Due to low image qualities, \ac{TIR} images show high uncertainty}
    \label{fig:cal_results2}
\end{figure}

We perform similar experiments with RGB and \ac{TIR} cameras. Due to the high noise in real images and the deviation from the ideal camera model, obtaining reliable calibration results with only six images is difficult. Therefore, we increase the number of image samples to 8 for RGB and 11 for \ac{TIR}\footnote{\ac{TIR} images possess high distortion and boundary blur due to thermal conduction, which requires more images for calibration}. Since there is no ground truth value of the camera parameters in the real world, we compare both the variance of estimated camera parameters and reprojection errors of whole images using the mean values of camera parameters.
The RGB camera used in the experiment is a \texttt{Blackfly-U3-16S2C-CS}  model from \texttt{FLIR} with a resolution of $2048 \times 1536$, while the \ac{TIR} camera is the \texttt{FLIR A65} model with a resolution of $640 \times 512$. The sample images and detection results are displayed in \figref{fig:cal_results2}.

\begin{table}[t!]
    \centering
    \caption{\textbf{Calibration results of real images.} }
    \label{tab:cal_results2}
    \resizebox{1.0\columnwidth}{!} {
    \renewcommand{\arraystretch}{1.2}
    \begin{tabular}{l|ccccc|c|c}    
        \hline
        \multicolumn{1}{c|}{Methods} & $\Delta f_x$ & $\Delta f_y$ &  $\Delta c_x$  & $\Delta c_y$ & $\Delta d_1$ & error &\#fails\\
        \hline
        \multicolumn{8}{c}{RGB (Random sampling 8 images from 16 images)}\\
        \hline
        B (baseline) &   $\pm $3214	&$\pm $18675	&$\pm $891	&$\pm $253&$\pm $13.1 & 30.77	&6\\
        B+D &   $\pm $15.4	& $\pm $15.7	&$\pm $16.2	&$\pm $13.5 &$\pm $0.009 & 1.68	&0\\
        B+D+E & $\pm $13.8	& $\pm $13.8	& $\pm $14.7	&$\pm $13.2&$\pm $0.009&1.66	&0\\
        B+D+E+O & $\pm$ \textbf{9.2} &$\pm$\textbf{9.5} &$\pm$\textbf{10.6} &$\pm$\textbf{9.0}&$\pm$\textbf{0.007} &\textbf{1.64} &0\\
        \hline
        \multicolumn{8}{c}{TIR (Random sampling 11 images from 16 images)}\\
        \hline
        B (baseline) &   $\pm $109	&$\pm $129	&$\pm $164	&$\pm $108&$\pm $1.942 & 6.33	&3\\
        B+D &   $\pm $44.5	& $\pm $46.4	&$\pm $45.1	&$\pm $8.5&$\pm $0.099 & 0.58	&0\\
        B+D+E & $\pm $44.6	& $\pm $46.5	& $\pm $44.8	&$\pm $8.1 &$\pm $0.089 &0.59	&0\\
        B+D+E+O & $\pm$ \textbf{1.2} &$\pm$\textbf{1.2} &$\pm$\textbf{0.5} & $\pm$\textbf{1.0}  & $\pm$\textbf{0.084} &\textbf{0.40} &0\\
        \hline
        \multicolumn{7}{l}{D: uncertainty-based detector, E: unbiased estimator, O: robust optimization}
    \end{tabular}
    \renewcommand{\arraystretch}{1}
    }
\end{table}

As a result, the potential of combining three modules (\ie, detector, estimator, optimizer) is prominent in real-world experiments. \tabref{tab:cal_results2} clearly demonstrates that the final version (B+D+E+O) has the lowest variance in estimated camera parameters and reprojection error. 
Interestingly, even though the final version uses a refined loss function as \eqoref{eq:robust_loss}, it shows a lower reprojection error than B+D+E, which directly minimizes the reprojection error itself.
This implies that our loss function is not only robust but also provides a better solution by avoiding local optima. The reprojection error of \ac{TIR} is lower than that of RGB because the reprojection error is proportional to image resolution, and we used more images for \ac{TIR} camera calibration.

%% file: sec/6_guide.tex
\section{Discussion on good camera calibration}
\label{sec:calibration_guide}
Despite using the same method, the calibration performance highly depends on individuals due to numerous hyperparameters. In this section, we discuss the meaning of each hyperparameter and how to set them for the good camera calibration.
The crucial hyperparameters (\ie the number of circles, the circle radius, the distance between circles, the number of total images, and the camera poses) are divided into two groups. The former three variables are related to the design of the target board, and the latter two variables are related to how we acquire images using the target. 

\subsection{How to design a good target} 
The more circles imply the more measurements. Since we estimate the pose of the target board based on the premise of planarity and a pre-known grid structure, a large number of control points enables the optimization to be robust against outliers. 
Next, the measurement quality is proportional to the circle radius. As the radius of the circle increases, the size of the blob observed in the image also increases. Bigger blobs include more boundary points, which contribute to low measurement uncertainty.
Lastly, the distance between circles is related to the coverage and detection difficulty.
As the distance between the circles increases, the area covered by the target in the image becomes larger, and the difficulty of circle detection decreases.

However, these three variables are not independent when the target board size is fixed. The total number of circles should decrease if we increase the circle radius or distance. Since the corner point of the checkerboard is a zero-dimensional feature, the size of the square does not affect the detection accuracy. Therefore, they increase the number of corner points as much as possible to overcome the error from imprecise measurements. Contrary to the checkerboard pattern, the circular pattern offers the flexibility to choose between increasing the number of points to enhance robustness or increasing the circle size to improve the measurement quality. In our experience, twelve circles with $4\times 3$ grid structure are sufficient to obtain accurate camera parameters. Asymmetric grids can also serve as an alternative and help prevent multiple solutions arising from rotational symmetry under 180-degree rotation. However, since the two solutions are sufficiently far apart in the $\ensuremath{\mathrm{se}(3)}$ manifold, there is no significant difference in accuracy. Therefore, we recommend choosing a target design considering manufacturing simplicity.
\begin{figure}[!t]
    \centering
    \includegraphics[trim=42 27 44 27, clip,width=0.99\columnwidth]{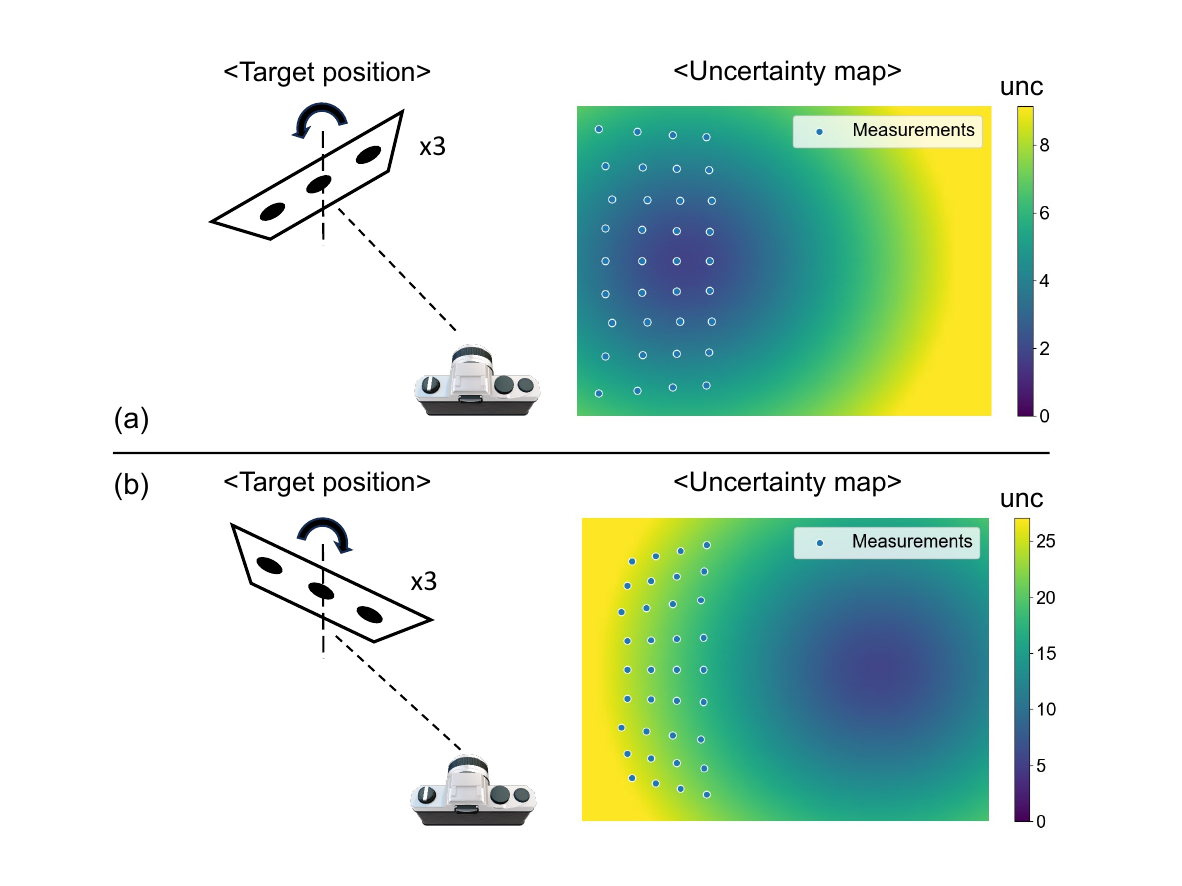}
    \caption{\textbf{Uncertainty map under opposite rotation.} The distribution of the uncertainty map follows the rotation of the target, not the translation or the measurement locations in the image. }
    \label{fig:guide_rot}
\end{figure}

\subsection{How to acquire good images}
The 3D relative pose (translation and rotation) between the camera and the target is more critical than the aforementioned hyperparameters. Several works~\cite{ICCV-2019-Peng, IROS-2013-richardson} address this by proposing a next-best-view strategy. However, that strategy depends on a pre-existing image set and does not specify how to select an optimal initial set for calibration. In practice, camera calibration typically involves capturing all images from scratch; therefore, it is preferable to acquire a globally optimal image set rather than iteratively choosing the next-best-view.

This section seeks the best target poses that result in a low uncertainty map. As illustrated in \figref{fig:calibration_uncertainty}, inappropriate target image sets decrease calibration performance and induce high uncertainty in certain regions of the image. Moreover, if two target poses satisfy the degeneracy condition, one image does not provide any information for calibration, making the optimization unstable or even fail. 

The common misconception is that each measurement lowers the uncertainty of its contiguous area. Translation is the dominant factor in deciding the measurement distribution in the image; however, the rotation of the target is more critical to determine the calibration uncertainty. See an example in \figref{fig:guide_rot}. The targets are located on the left side of the camera; therefore, the measurements of both cases (a) and (b) are distributed on the left side of the images. Nevertheless, the uncertainty maps of the two cases show the opposite tendency. The first case (a) is aligned with the common intuition, but the second case lowers right-side uncertainty, which contradicts the premise. The self-rotation direction of the target, not the measurement locations, decides the region where the information from the target image affects the uncertainty map.

The reason for this phenomenon is based on the degeneracy condition. According to the literature \cite{TPAMI-2000-zhang}, targets satisfying the coplanar condition cause degeneracy in camera calibration. It means that moving the target in an orthogonal direction to the normal vector of the target plane does not provide additional information. Therefore, they recommend acquiring target images in diverse rotation angles. Applying this principle to the calibration uncertainty makes it straightforward that the normal vector of the target plane decides the region where the uncertainty will decrease. The location of measurements is freely adjusted by moving the target in an orthogonal direction. Hence, acquiring images from diverse rotations is essential.

\begin{figure}[!t]
    \centering
    \includegraphics[trim=19 28 19 21, clip,width=1.0\columnwidth]{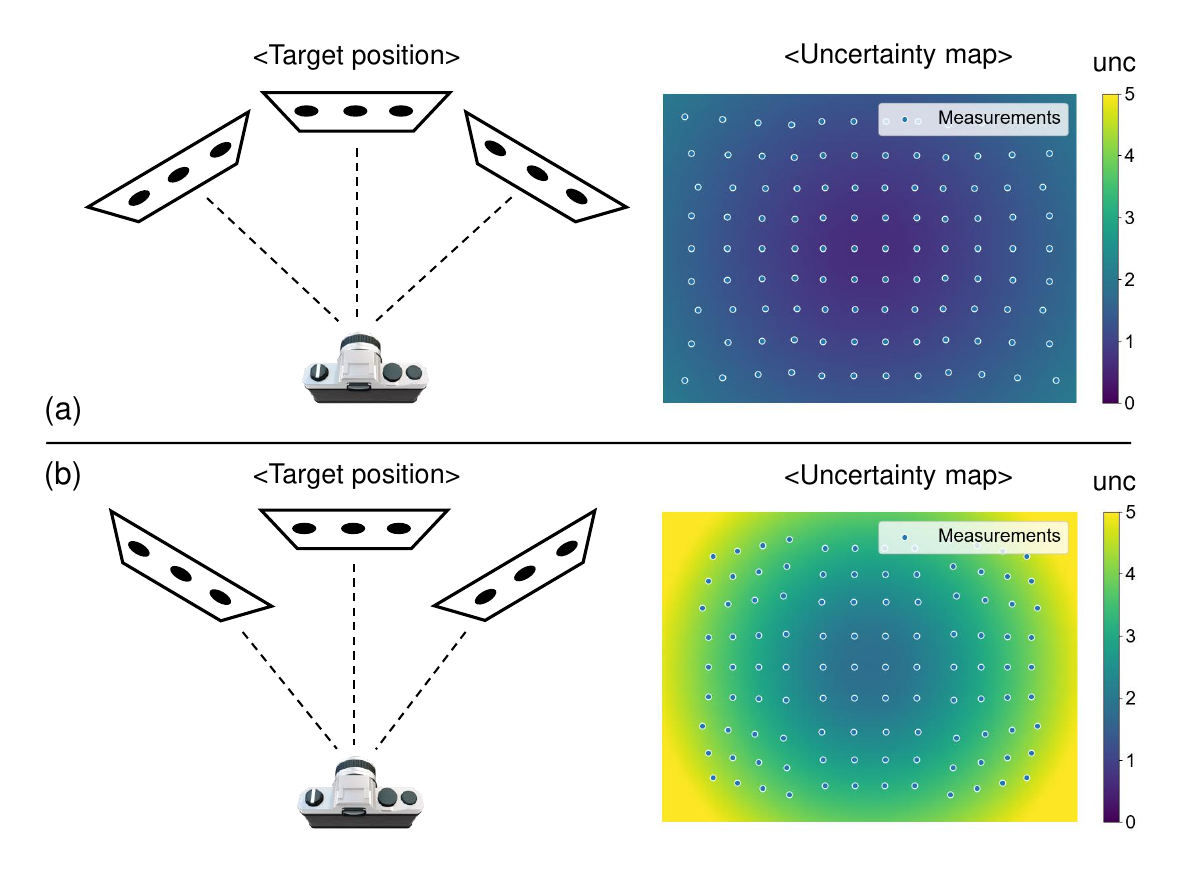}
    \caption{\textbf{Uncertainty variation due to the combination of rotation and translation.} The two cases have the same translation and rotation candidates, but the combination is different. The revolution and rotation direction are equal in (a), and opposite in (b). As a result, they show a substantial difference in overall uncertainty.}
    \label{fig:uniform_cases}
\end{figure}

\begin{figure}[!t]
    \centering
    \includegraphics[trim=20 23 35 13, clip,width=1.0\columnwidth]{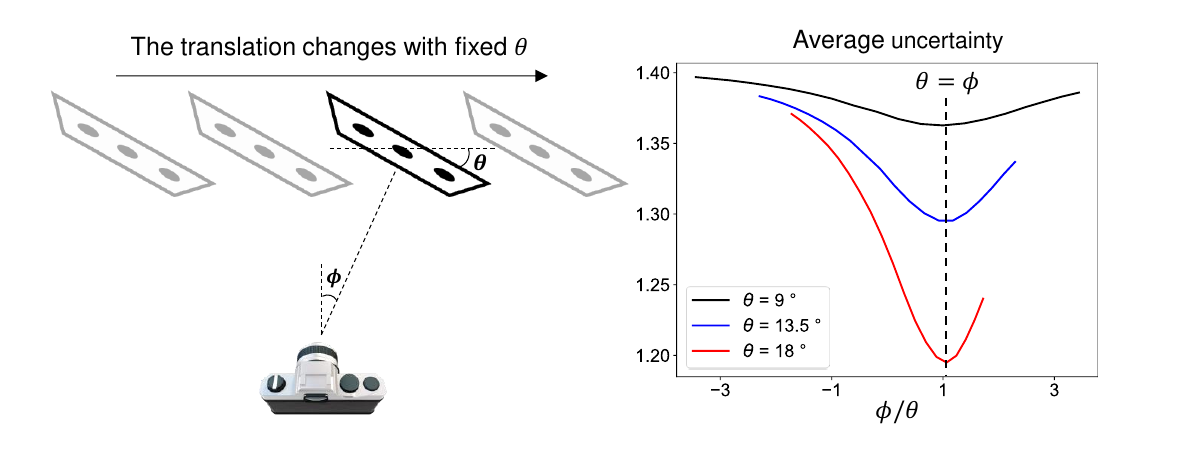}
    \caption{\textbf{Optimal translation for a given rotation.} We investigate the changes in the overall uncertainty while moving the target from the left side to the right side. Interestingly, when the revolution angle $\phi$ is equal to the rotation angle $\theta$, the uncertainty is minimized regardless of the absolute value of $\theta$.}
    \label{fig:guide_trans}
\end{figure}

Nonetheless, the translation also plays a crucial role in the camera calibration. \figref{fig:uniform_cases} illustrates uncertainty maps of two cases. The first case (a) is when the rotation direction of the target aligns with the revolution direction of the target according to the camera. The second case (b) is the opposite. These two cases have the same rotation candidates, but the combination of rotations and translations is different. As a result, their uncertainty maps significantly differ, while case (a) definitely shows lower uncertainty than case (b). This implies that although the rotation value determines which image region the information is provided in, the extent of this depends on the translation value. This principle is also observed in the previous figure. The maximum uncertainty of \figref{fig:guide_rot} (b) is four times higher than (a).

The remaining question is what the optimal translation is for a given rotation of the target board. To find the answer, we move the target board from the left side to the right side while maintaining the rotation value of the target board as $\theta$. Then, we investigate when the average uncertainty (\ie the spatial average of the scalar uncertainty in the uncertainty map) becomes the lowest\footnote{Since it is impossible to conduct calibration with one image, we add six fixed images, ensuring that the target's movement path does not overlap with these images.}.
\figref{fig:guide_trans} describes the experimental conditions and the result. The optimal translation value varies according to the target rotation value; however, surprisingly, we found that the ratio between the rotation $\theta$ and the optimal revolution value $\phi$ is always one. It means that when the vector from the camera to the target center is parallel to the target normal, the overall uncertainty is minimized. The results of this experiment explain why case (b) shows higher uncertainty than case (a) in the previous figures: \figref{fig:guide_rot} and \figref{fig:uniform_cases}. Another advantage of this condition is that circles on the target board are observed as large as possible, which means the measurement uncertainty is minimized at this configuration.

\begin{figure*}[!t]
    \centering
    \includegraphics[trim=30 27 10 30, clip,width=0.9\textwidth]{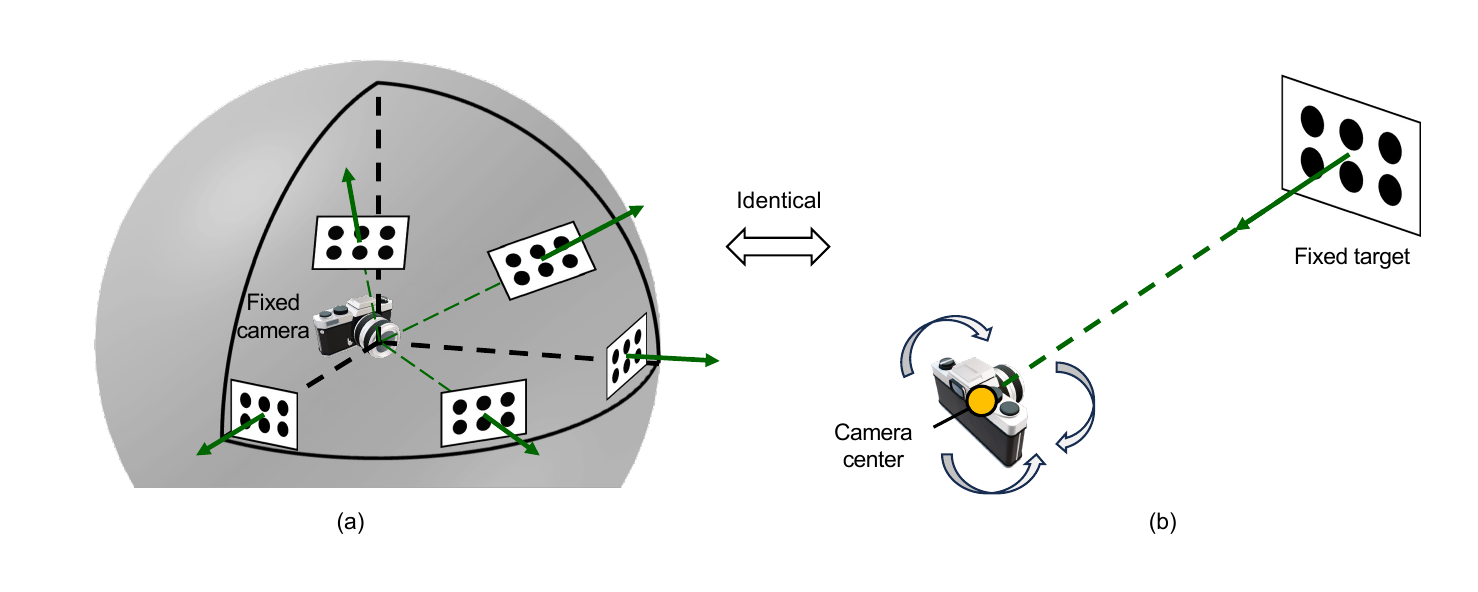}
    \caption{\textbf{Optimal configurations.} (a) \textbf{Fixed-camera version}. To perform optimal camera calibration, the target board should be located on the tangential planes of an imaginary sphere centered around the camera. (b) \textbf{Fixed-target version}. The optimal configuration is also achieved by placing the camera center along the extension of the target board’s z-axis and applying rotation only. Those two configurations are identical.}
    \label{fig:guide_optimal}
\end{figure*}



To encapsulate the above discussion, target images should be obtained in various rotations to reduce overall uncertainty equally in the image domain, while preserving the $\theta = \phi$ condition mentioned earlier to maximize efficacy. An easy way to satisfy both conditions simultaneously is to consider an imaginary sphere centered around the camera. By capturing images from various locations while keeping the target tangentially in contact with the sphere, as shown in \figref{fig:guide_optimal} (a), optimal calibration can be achieved. An alternative way to satisfy this condition is to locate the camera directly in front of the target board (\ie the camera center is located along the extension of the target’s z-axis) and apply rotation only to the camera (not the target) as \figref{fig:guide_optimal} (b). This approach is useful when the target is fixed and immovable. Those two approaches are completely identical from the perspective of the relative configuration between the camera and the target.

The radius of the sphere (not circle radius) is meaningless in the projection geometry due to the homogenous property (scale is ignored). However, measurements are highly affected by the sphere radius. A large sphere radius implies a greater distance between the camera and the target, which causes higher measurement uncertainty. Nevertheless, far-distance images are required for high rotation angles. As the rotation value increases, the target’s position shifts toward the periphery of the camera’s field of view, thereby increasing the probability that the circular pattern will fall outside the image area. In this case, increasing the sphere radius reduces the circle sizes and allows the circles to be projected inside the image area. 

To sum up, we recommend following the process below. 
\begin{itemize}
    \item maintaining the target board being tangential to the imaginary sphere centered around the camera
    \item acquire low rotation in a small radius (close distance)
    \item acquire high rotation in a large radius (far distance)
\end{itemize}
It is important to capture images uniformly from all rotations, and a total of 16 images—8 from close range and 8 from a farther distance—is sufficient. We provide a sample image set in \bl{Appendix E}.


%% file: sec/7_conclusion.tex
\section{conclusion}
This paper is the first to propose a method for integrating the concept of uncertainty into camera calibration using the circular pattern. The centroid uncertainty of a two-dimensional shape is derived from mathematical thoroughness and investigated in diverse circumstances. By applying this uncertainty in each calibration process, including detection and optimization, the calibration accuracy and robustness increase significantly. With our unbiased estimator \cite{CVPR-2024-song}, we have finally completed the calibration framework via the circular pattern. Furthermore, we introduce the principles of the 6D configuration between the camera and the target. It provides guidelines for non-experts to achieve consistent and reliable calibration results.

The circular pattern, with our framework, demonstrates significantly better performance compared to existing methods. Especially, in special vision sensors with low image quality, its advantages are maximized. Our method is not limited to intrinsic camera calibration, but could be widely utilized in 3D pose estimation, such as extrinsic calibration between multimodal sensors or camera pose estimation using the circular pattern.

%% file: sec/8_biography.tex
\newpage
\section{Biography Section}
\begin{IEEEbiography}[{\includegraphics[width=1in,height=1.25in,clip,keepaspectratio]{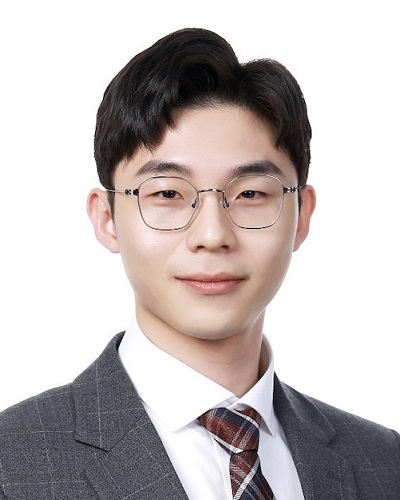}}]{Chaehyeon Song}
received the B.S. degree in 2022 and the M.S. degree in 2024 from Seoul National University, Seoul, South Korea, respectively, all in mechanical engineering. His research area is 3D vision, Geometry, and learning-based perception.
\end{IEEEbiography}
\begin{IEEEbiography}[{\includegraphics[width=1in,height=1.25in,clip,keepaspectratio]{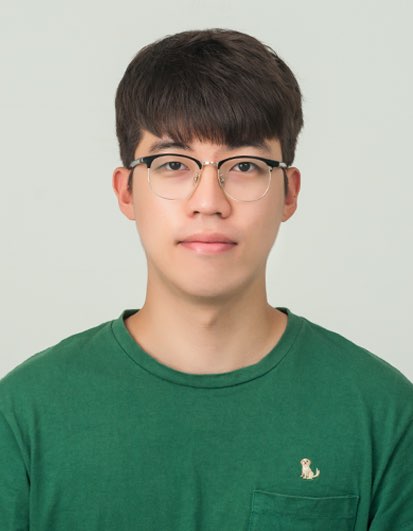}}]{Dongjae Lee}
received the B.S. degree in mechanical engineering from Seoul National University (SNU), Seoul, South Korea, in 2023. He is currently pursuing the Ph.D. degree in mechanical engineering at SNU. His research interests include robotic perception, with a focus on LiDAR-based localization, mapping, and long-term scene understanding in dynamic environments.
\end{IEEEbiography}
\begin{IEEEbiography}[{\includegraphics[width=1in,height=1.25in,clip,keepaspectratio]{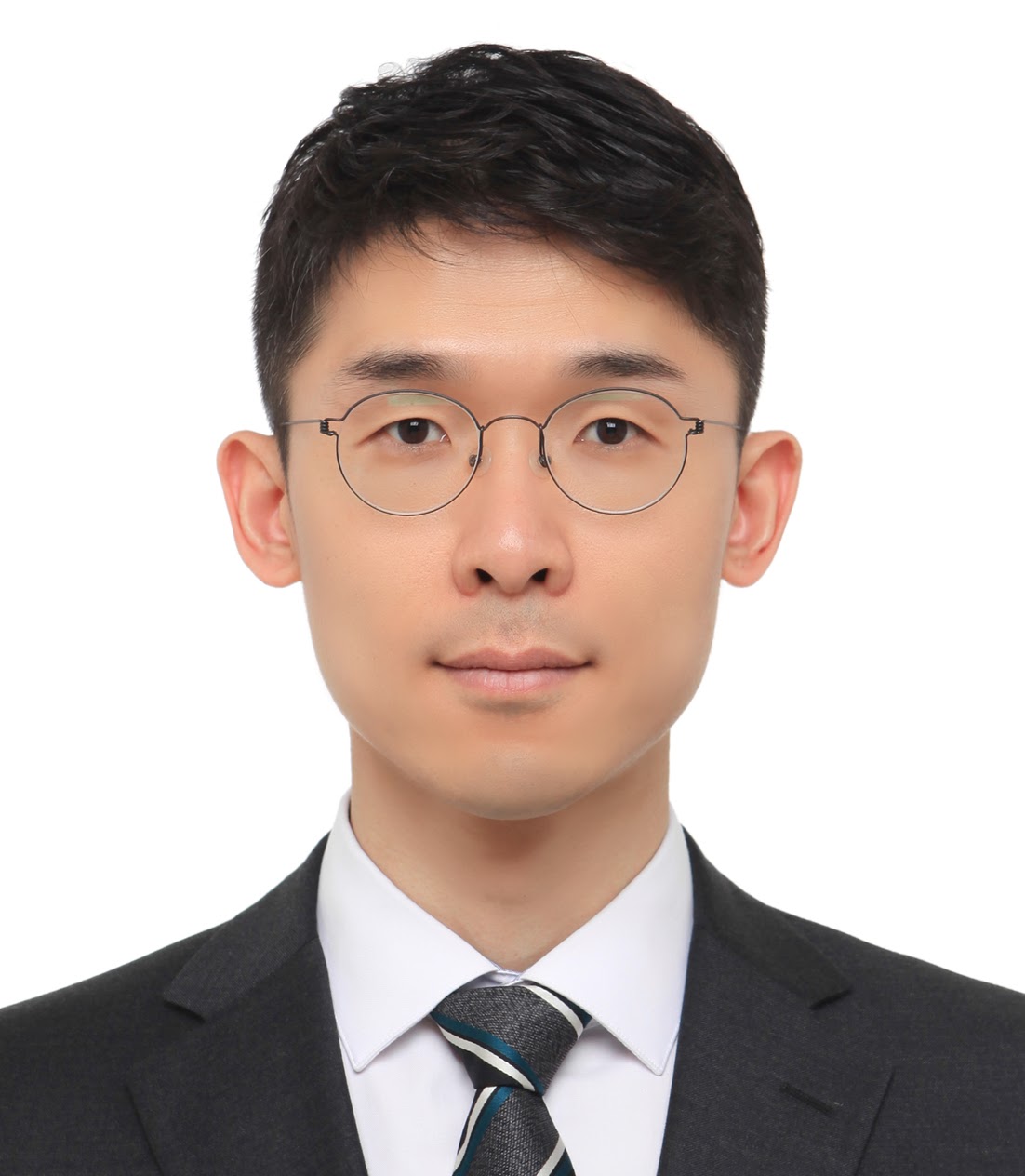}}]{Jongwoo Lim}
Jongwoo Lim received the BS degree from Seoul National University, Seoul, Korea, in 1997, and the MS and PhD degrees from the University of Illinois, Urbana-Champaign, in 2003 and 2005, respectively. He was at Honda Research Institute Inc., Mountain View, CA, as a senior scientist from 2005 to 2011, and at Google Inc., Mountain View, CA, as a software engineer from 2011 to 2012. He was an associate professor with the Department of Computer Science, Hanyang University, Seoul, Korea. Currently, he is a professor with the Department of Mechanical Engineering, Seoul National University, Seoul, Korea. His research interests include computer vision, robotics, and machine learning.
\end{IEEEbiography}
\begin{IEEEbiography}[{\includegraphics[width=1in,height=1.25in,clip,keepaspectratio]{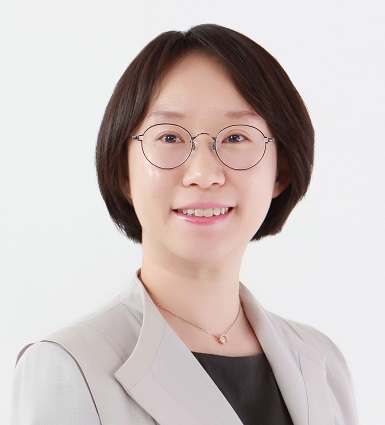}}]{Ayoung Kim}%
(S'08--M'13--S'23) received the B.S. and M.S. degrees from Seoul National University (SNU) in 2005 and 2007, and the Ph.D. degree from the University of Michigan (UM), Ann Arbor 2012. She was an associate professor at Korea Advanced Institute of Science and Technology (KAIST) from 2014 to 2021. Currently, she is an associate professor at SNU.
\end{IEEEbiography}%

 



%% file: sec/9_appendix.tex
{\appendices
\newpage
\onecolumn

\normalsize
\section{Proof of \lemref{lemma:info_full}}
\label{ap:lemma_info_full}
Since $\bs{X}$ is fully connected, the eigenvalues of the covariance should be $\lambda_1>>1$ and $\lambda_2 = \lambda_3 =\cdots = \lambda_n$, while the corresponding eigenvector $\bs{v}_1 = [1,1,\cdots,1]^{\top}/\sqrt{n}$. The following covariance satisfies the above conditions.
\begin{equation}
    \bs{\Sigma} = \begin{bmatrix}
        t+s & t & \cdots & t\\
        t & t+s & \cdots & t\\
        \multicolumn{4}{c}{\vdots}\\
        t & t & \cdots & t+s\\
    \end{bmatrix},
\end{equation}
whose eigen values are $\{s+nt, s, \cdots, s\}$. Its inverse, the information matrix, is derived as
\begin{equation}
    \bs{\Omega}= \frac{t}{(nt+s)s}\begin{bmatrix}
        n-1 & -1 & \cdots & -1\\
    -1 & n-1 &  \cdots & -1\\
    \multicolumn{4}{c}{\vdots}\\
    -1 & -1 & \cdots & n-1\\
    \end{bmatrix} + \frac{1}{nt+s}\bs{I},
\end{equation}
$p(u_i) \sim N(0, \infty)$ condition can be satisfied by taking the limit of $t$ as it approaches infinity. Then, $\bs{\Omega}$ converge to 
\begin{equation}
    \lim_{t \to \infty}\bs{\Omega}= \frac{1}{ns}\begin{bmatrix}
        n-1 & -1 & \cdots & -1\\
    -1 & n-1 &  \cdots & -1\\
    \multicolumn{4}{c}{\vdots}\\
    -1 & -1 & \cdots & n-1\\
    \end{bmatrix}.
\end{equation}
The normalization factor $s$ is obtained from the condition $p(u_i|u_j) \sim N(0,\sigma^2)$.
\begin{align}
     \bs{\Sigma}(u_i|u_j) &= \bs{\Sigma}_{ii}-\bs{\Sigma}_{ij}\bs{\Sigma}_{jj}^{-1}\bs{\Sigma}_{ji},\\
     &=  t+s - \frac{t^2}{t+s}= \frac{2ts+s^2}{t+s} = \sigma^2.  \\
     \lim_{t \to \infty}\bs{\Sigma}(u_i|u_j) &= 2s =\sigma^2 ,\\
     \therefore \lim_{t \to \infty} s &= \sigma^2/2.
\end{align}

\section{Normalization factor}
\label{ap:z}
Let matrix $\bs{A}_n$ is a $n\times n$ tridiagonal matrix as
\begin{equation}
    \bs{A}_n= \begin{bmatrix}
        2 & -1 &0&0& \cdots & 0&0\\
        -1& 2 &  -1&0& \cdots &0& 0\\
        0 & -1 & 2 &-1& \cdots &0& 0 \\
        0 & 0 & -1 &2& \cdots &0& 0 \\
        \multicolumn{7}{c}{\vdots}\\
        0 & 0 &0&0& \cdots & 2&-1\\
        0 & 0 &0&0& \cdots & -1&2\\
    \end{bmatrix}.
\end{equation}
Note that $\bs{A}_n$ is tridigonal and differs from $\bs{\Omega}_{prior\_x}$. Let $a_n = det(\bs{A}_n)$. Then $a_n$ satisfies the following.
\begin{align}
    a_n &= 2*a_{n-1} + a_{n-2}\\
    \longleftrightarrow a_n & = c_1(1+\sqrt{2})^{n} + c_2(1-\sqrt{2})^n ,\\
    \label{eq:an}
    \therefore \lim_{n \to \infty} \frac{a_{n-1}}{a_n} &= \frac{1}{1+\sqrt{2}}=\sqrt{2}-1 .
\end{align}

From \lemref{lemma:information} and \eqoref{eq:prior},\
\begin{equation}
    \bs{\Omega}_{prior\_x}({u}_1,\cdots, {u}_{n-1}|{u}_{n}) =\frac{1}{z\sigma^2}\bs{A}_{n-1}.
\end{equation}
W.L.O.G., set $i$ to one. Then,
\begin{align}
    \frac{1}{\sigma^2}&=\bs{\Omega}_{prior\_x}({u}_i|{u}_{i+1})=\bs{\Omega}_{prior\_x}({u}_1|{u}_{n}) ,\\
    &=\frac{1}{z\sigma^2}\bs{A}_{n-1}(0,0) - \bs{A}_{n-1}(0,:)\bs{A}_{n-2}^{-1}\bs{A}_{n-1}(:,0), \hspace{1cm}(\text{by \lemref{lemma:information}})\\
    &=\frac{1}{z\sigma^2}\left( 2 - \begin{bmatrix}
        -1 & 0&\cdots & 0
    \end{bmatrix}\bs{A}_{n-2}^{-1}
    \begin{bmatrix}
        -1\\
        0\\
        \vdots\\
        0
    \end{bmatrix}\right),\\
    &= \frac{1}{z\sigma^2}\left(2-\frac{a_{n-3}}{a_{n-2}} \right),\\
    \therefore \hspace{1mm} z(n) &=(2-\frac{a_{n-3}}{a_{n-2}}).
\end{align} 
By \eqoref{eq:an},
\begin{equation}
    \lim_{n \to \infty} z(n) =\lim_{n \to \infty} 2-\frac{a_{n-3}}{a_{n-2}} = 3-\sqrt{2}.
\end{equation}

\vspace{1cm}
\section{Jacobian tendency}
\label{ap:jacob_tendency}
\begin{equation}
    (\bar{u})_i' = \frac{1}{M^{00}} \begin{bmatrix}
        -\left(\frac{u_{i+1}+2u_i}{3} - \bar{u} \right)\frac{\Delta v_i^{i+1}}{2} - \left(\frac{2u_{i}+u_{i-1}}{3} - \bar{u} \right)\frac{\Delta v_{i-1}^{i}}{2}\\
        \left( \frac{u_{i+1}+u_{i}+u_{i-1}}{3}-\bar{u} \right) \frac{\Delta u_{i-1}^{i+1}}{2}
    \end{bmatrix}^T
\end{equation}
1. If $u_i$ is far to $\bar{u}$ ,  $(\bar{u})_i'$ is orthogonal to contour\\
\begin{equation}
    u_{i-1}-\bar{u} \simeq u_i -\bar{u}\simeq u_{i+1}-\bar{u},
\end{equation}
\begin{equation}
    (\bar{u})_i' \simeq \frac{u_i - \bar{u}}{2M^{00}} \begin{bmatrix}
        -(v_{i+1}-v_{i-1})\\ (u_{i+1}-u_{i-1})
    \end{bmatrix}^T \bot \begin{bmatrix}
        \Delta u_{i-1}^{i+1}\\ \Delta v^{i+1}_{v_i} 
    \end{bmatrix}^T ,
\end{equation}
where, vector $[\Delta u_{i-1}^{i+1} ,\Delta v^{i+1}_{v_i} ]$ indicate the tangential direction of the contour at $(u_i, v_i)$.\\

\noindent 2. if $u_i$ is close to $\bar{u}$, the norm of  $(\bar{u})_i' \simeq 0$ 
\begin{align}
    |\Delta u_{t}^{t+1}| , |\Delta v_t^{t+1}|  &< 1 \text{ for } \forall t  \\
    \left|\frac{u_{i+1}+2u_i}{3} - \bar{u} \right| &< |u_i - \bar{u}| + 1/3  \\
    \left|\frac{2x_{i}+u_{i-1}}{3} - \bar{u} \right| &< |u_i - \bar{u}| + 1/3 \\
    \left| \frac{u_{i+1}+u_{i}+u_{i-1}}{3}-\bar{u} \right| &< |u_i - \bar{u}| + 1/3 \\
\end{align}
\begin{equation}
    \therefore ||(\bar{u})_i'|| \le \frac{\sqrt{2}/3+\sqrt{2}|u_i-\bar{u}|}{M^{00}} \simeq 0 \quad (M^{00} >> |u_i - \bar{u}|)
\end{equation}

\newpage
\section{Invariance and Equivariance}
\label{ap:invariance}
Let $\Tilde{\bs{x}}' = \bs{R}\Tilde{\bs{x}} + \bs{t}$ and $I'(\Tilde{\bs{x}}') = I(\Tilde{\bs{x}})$. Then,
\begin{align}
    \Tilde{\bs{X}}' &= \begin{bmatrix}
        \bs{R} & \bs{0} & \cdots & \bs{0}\\
        \bs{0} & \bs{R} & \cdots & \bs{0}\\
        \multicolumn{4}{c}{\vdots} \\
        \bs{0} & \bs{0} & \cdots & \bs{R}
    \end{bmatrix}\Tilde{\bs{X}} + \begin{bmatrix}
        \bs{t} & \bs{0} & \cdots & \bs{0}\\
        \bs{0} & \bs{t} & \cdots & \bs{0}\\
        \multicolumn{4}{c}{\vdots} \\
        \bs{0} & \bs{0} & \cdots & \bs{t}
    \end{bmatrix},\\
    & = \overleftrightarrow{\bs{R}}\Tilde{\bs{X}} + \overleftrightarrow{\bs{t}} .
\end{align}
Let $\bs{A}_{ij} $  is $2n \times 2n $ matrix whose elements are zero except for
\begin{equation}
    \label{eq:temp1}
    \bs{A}_{ij}[2i:2i+4, 2j:2j+4] = \begin{bmatrix}
             \bs{I}_2  & -\bs{I}_2\\
             -\bs{I}_2 & \bs{I}_2
         \end{bmatrix},
\end{equation}
where $\bs{I}_2$ is $2\times2$ identity matrix and
\begin{equation}
    \label{eq:temp2}
    \begin{bmatrix}
        \bs{R} & \bs{0}\\
        \bs{0} & \bs{R}
    \end{bmatrix}\begin{bmatrix}
            \bs{I}_2  & -\bs{I}_2\\
             -\bs{I}_2 & \bs{I}_2
         \end{bmatrix}\begin{bmatrix}
        \bs{R}^{\top} & \bs{0}\\
        \bs{0} & \bs{R}^{\top}
    \end{bmatrix}= \begin{bmatrix}
             \bs{I}_2  & -\bs{I}_2\\
             -\bs{I}_2 & \bs{I}_2
         \end{bmatrix}.
\end{equation}
By \eqtref{eq:temp1}{eq:temp2}, $\bs{\Omega}_{prior}$ is rotation invariant
\begin{align}
    \dvec{\bs{R}}  \bs{\Omega_{prior}}\dvec{\bs{R}}^{\top}  & = \frac{1}{z\sigma^2}\dvec{\bs{R}} \left( \sum_{i=1}^n \bs{A}_{ij} \right) \dvec{\bs{R}}^{\top},\\
     & = \frac{1}{z\sigma^2} \sum_{i=1}^n \dvec{\bs{R}}\bs{A}_{ij} \dvec{\bs{R}}^{\top},\\
     & = \frac{1}{z\sigma^2}\sum_{i=1}^n\bs{A}_{ij},\\
     \label{eq:temp4}
     & = \bs{\Omega_{prior}} .
\end{align}
The likelihood term is transformed as follows.
\begin{align}
    \nabla I'(\Tilde{\bs{x}}') &=  \bs{R}\nabla I(\Tilde{\bs{x}}),\\
    \label{eq:temp3}
    \therefore \bs{\Omega}_i' &= \bs{R}\bs{\Omega}_i\bs{R}^{\top}.
\end{align}
Combining \eqtref{eq:temp4}{eq:temp3}, we get
\begin{align}
    \bs{\Omega}' &= \bs{\Omega_{prior} + \begin{bmatrix}
        \bs{R}\bs{\Omega}_1\bs{R}^{\top} & \bs{0} & \cdots & \bs{0}\\
        \bs{0} & \bs{R}\bs{\Omega}_2\bs{R}^{\top} & \cdots & \bs{0}\\
        \multicolumn{4}{c}{\vdots}\\
        \bs{0} & \bs{0} & \cdots & \bs{R}\bs{\Omega}_n\bs{R}^{\top}\\
    \end{bmatrix}},\\
    &= \dvec{\bs{R}}  \bs{\Omega_{prior}}\dvec{\bs{R}}^{\top} + \dvec{\bs{R}}  \begin{bmatrix}
        \bs{\Omega}_1 & \bs{0} & \cdots & \bs{0}\\
        \bs{0} & \bs{\Omega}_2 & \cdots & \bs{0}\\
        \multicolumn{4}{c}{\vdots}\\
        \bs{0} & \bs{0} & \cdots & \bs{\Omega}_n\\
    \end{bmatrix}\dvec{\bs{R}}^{\top},\\
    &= \dvec{\bs{R}}  \left(\bs{\Omega_{prior}}+\begin{bmatrix}
        \bs{\Omega}_1 & \bs{0} & \cdots & \bs{0}\\
        \bs{0} & \bs{\Omega}_2 & \cdots & \bs{0}\\
        \multicolumn{4}{c}{\vdots}\\
        \bs{0} & \bs{0} & \cdots & \bs{\Omega}_n\\
    \end{bmatrix} \right)\dvec{\bs{R}}^{\top},\\
    \label{eq:info_invariant}
    &=\dvec{\bs{R}}  \bs{\Omega}\dvec{\bs{R}}^{\top}.
\end{align}

\eqoref{eq:info_invariant} implies our shape uncertainty is translation invariant and rotation equivariant.

\newpage
\section{Image sample for good calibration}
\label{ap:sample}
\begin{figure}[!h]
    \centering
    \includegraphics[trim= 20 10 20 10, clip,width=1.0\textwidth]{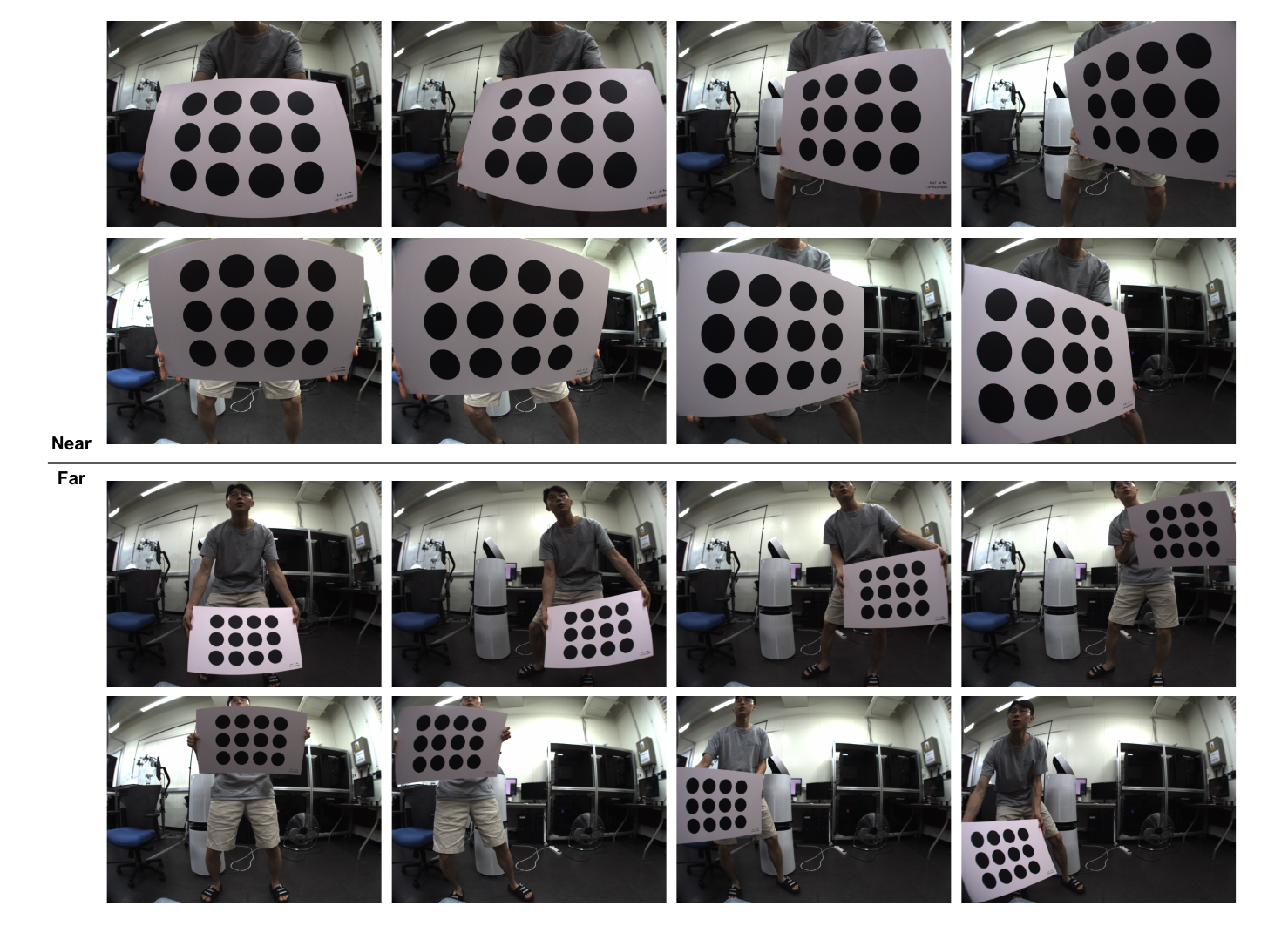}
    \caption{\textbf{Recommended configurations between the camera and the target for camera calibration} }
    \label{fig:ap_sample}
\end{figure}

}